\documentclass[lettersize,journal]{IEEEtran}
\usepackage{amsmath,amsfonts}
\usepackage{array}
\usepackage[caption=false,font=normalsize,labelfont=sf,textfont=sf]{subfig}
\usepackage{textcomp}
\usepackage{stfloats}
\usepackage{url}
\usepackage{verbatim}
\usepackage{graphicx}
\usepackage{cite}
\usepackage{amsthm}
\usepackage{amsmath}
\usepackage{multirow}
\usepackage{mathtools}
\usepackage{booktabs}
\usepackage{fontawesome5}
\usepackage[table]{xcolor}
\usepackage{color}
\usepackage{pifont}
\usepackage{wrapfig}
\usepackage{bbding}
\usepackage{algorithm}
\usepackage{algpseudocode}
\usepackage{amssymb} 
\usepackage[table]{xcolor}  

\usepackage{booktabs}
\usepackage{multirow}
\usepackage{threeparttable}
\usepackage{tabularx}
\newcolumntype{C}{>{\centering\arraybackslash}X}
\usepackage{enumitem}

\begin{document}

\title{PhyMRI-SR: Toward Physics-Aware MRI Image Super-Resolution}

\author{
Lihua Wei,
Huatong Gao,
Jia Gong,
Zhiyu Tan,
Hao Li,
Jun Liu,
and Zhihua Ren%
\thanks{Lihua Wei is with the School of Biomedical Engineering, ShanghaiTech University, Shanghai 201210, China.}%
\thanks{Huatong Gao is with the School of Biomedical Engineering, ShanghaiTech University, Shanghai 201210, China, and also with the Shanghai Academy of AI for Science, Shanghai, China.}%
\thanks{Jia Gong is with the Shanghai Academy of AI for Science, Shanghai, China.}%
\thanks{Zhiyu Tan and Hao Li are with the Shanghai Academy of AI for Science, Shanghai, China, and also with Fudan University, Shanghai, China.}%
\thanks{Jun Liu is with the School of Computing and Communications, Lancaster University, Lancaster, U.K.}%
\thanks{Zhihua Ren is with the School of Biomedical Engineering \& State Key Laboratory of Advanced Medical Materials and Devices, ShanghaiTech University, Shanghai 201210, China, and also with the Shanghai Clinical Research and Trial Center, Shanghai 200231, China.}%
\thanks{Lihua Wei, Huatong Gao and Jia Gong contributed equally to this work. Corresponding authors: Jia Gong and Zhihua Ren (e-mail: gongjia@sais.com.cn; renzhih@shanghaitech.edu.cn).}%
}

\markboth{Journal of \LaTeX\ Class Files,~Vol.~14, No.~8, August~2021}%
{Shell \MakeLowercase{\textit{et al.}}: A Sample Article Using IEEEtran.cls for IEEE Journals}

\IEEEpubid{0000--0000/00\$00.00~\copyright~2021 IEEE}

\maketitle

\begin{abstract}
Magnetic resonance imaging (MRI) super-resolution is vital for improving
diagnostic accessibility, yet most methods treat it as a deterministic
mapping from a fixed low-resolution input to a high-resolution target.
This overlooks a key property of MRI acquisition physics: spatial
resolution and signal-to-noise ratio (SNR) are inherently coupled, making
any given low-resolution scan merely one of many possible realizations
under varying acquisition trade-offs. We rethink MRI super-resolution as
a \emph{physics-aware reconstruction problem}, in which the goal is to
identify the optimal resolution--SNR configuration and then super-resolve
it to obtain high-quality MRI results. A key implication of this
formulation is that MRI resolution becomes dynamic rather than fixed. To
handle such resolution-heterogeneous inputs, we adapt 2D Gaussian
Splatting (2D GS) to MRI by formulating reconstruction as a
coordinate-based, resolution-agnostic rendering problem. To futher enhance the fidelity of MRI results, we introduce three innovations:
(1)~a \emph{prior-aware Gaussian representation} that combines an
Anatomical Structure Prior for tissue-specific kernel initialization with
an Imaging System Prior that captures hardware characteristics via a
covariance dictionary; (2)~a \emph{physics-constrained signal modeling}
scheme that predicts intrinsic tissue parameters (proton density $\rho$
and effective relaxation rate $R_2$) and synthesizes intensities through
governing physical equations, ensuring biophysically plausible contrast;
and (3)~a \emph{meta-learning framework} that alleviates paired-data
scarcity by pretraining on simulated data and adapting to real-world
conditions. Extensive experiments on dynamic-resolution datasets and normal benchmark demonstrate that our method achieves state-of-the-art
performance, highlighting its strong potential for clinical deployment. \textbf{Project Page:}
\url{https://bio-med-i2-lab.github.io/projects/PhyMRI-SR}.
\end{abstract}


\section{Introduction}
\label{sec:intro}

Magnetic resonance imaging (MRI) is a cornerstone of modern medical diagnostics, providing non-invasive, high-contrast visualization of soft tissues essential for detecting tumors, cerebrovascular diseases, and neurodegenerative disorders~\cite{kransdorf1994magnetic,detre1998noninvasive,islam2018brain}. However, acquiring high-quality MRI scans typically requires expensive hardware and prolonged acquisition times~\cite{heckel2024deep,zeng2021review}, significantly limiting accessibility. A promising alternative is to capture low-resolution images using cost-effective hardware and computationally reconstruct high-quality counterparts through super-resolution (SR)~\cite{van2012super,shi2018super,qiu2023medical}.

Early MRI super-resolution works~\cite{muhammad2024brain,yang2016sparse,plenge2012super} typically relied on handcrafted priors and precise image registration to model the correlation between high-resolution (HR) and low-resolution (LR) images, such as frequency-domain regularization~\cite{mayer2007measuring,tieng2011mri,luo2017fast} and spatial-domain interpolation~\cite{manjon2010mri,carmi2006resolution}.
However, relying heavily on handcrafted priors limited their ability to faithfully recover fine anatomical structures and made them sensitive to noise and motion artifacts~\cite{greenspan2002mri,freeman2002example,van2012super}. 
To address these limitations, recent approaches leverage deep learning to learn data-driven mappings from low- to high-resolution MRI images via CNNs~\cite{pham2019multiscale,zhou2022super,zhu2021residual,lyon2023spatio} or Transformers~\cite{lyu2023multicontrast,huang2023accurate,li2023rethinking}. In parallel, generative models~\cite{zhang2022soup,zhang2025super,wang2023inversesr,safari2025mri} have been explored to enhance structural realism by optimizing models to generate realistic MRI images.

\begin{figure}[t]
    \centering
    \includegraphics[width=\linewidth]{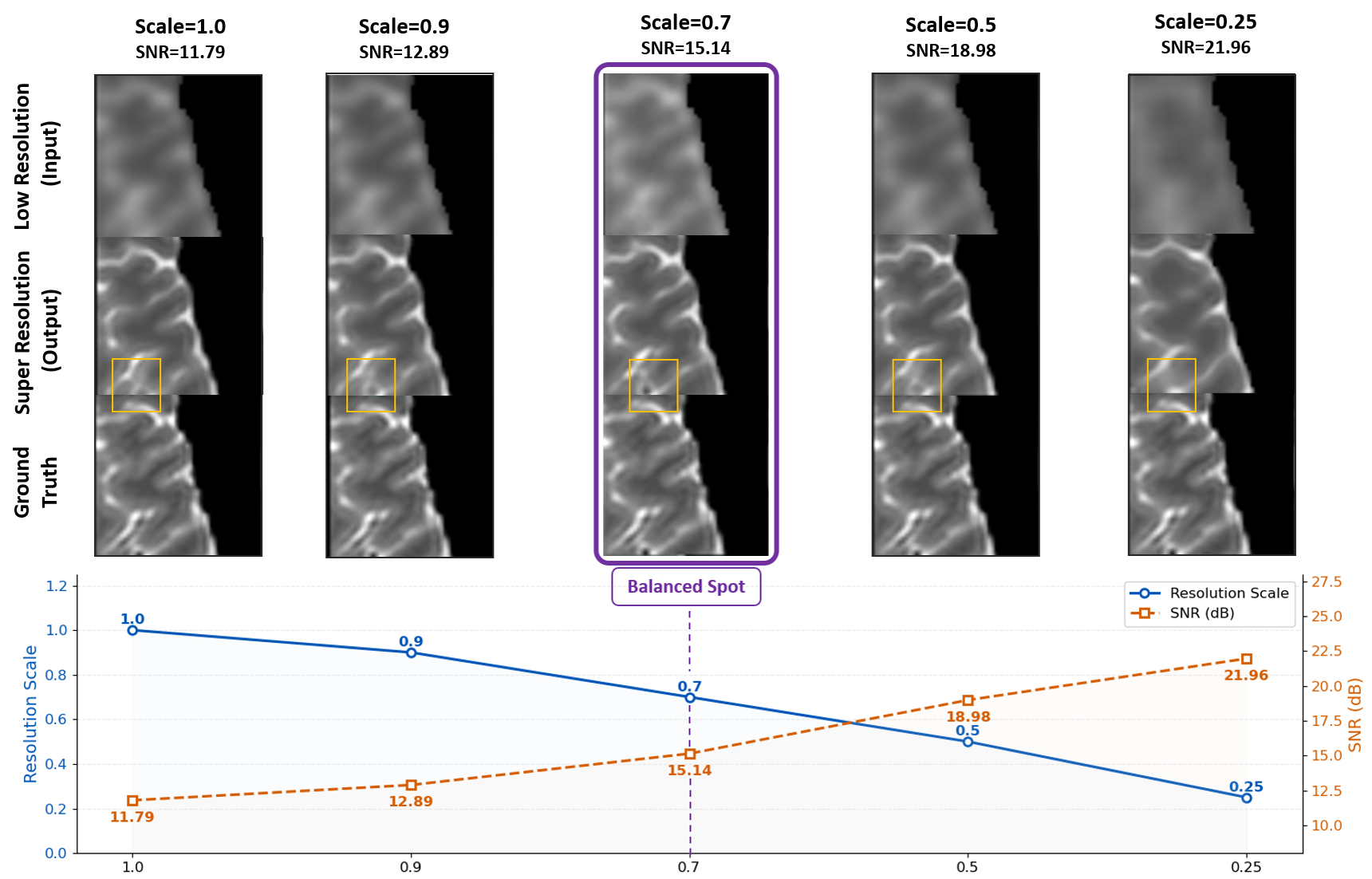}
    \caption{
    Illustration of the trade-off between spatial resolution and signal-to-noise ratio (SNR) under a simulated ultra-low MRI system (64 mT). 
    In the high-resolution but low-SNR setting, severe noise leads to fragmented and discontinuous anatomical structures, as highlighted in the yellow boxes. 
    The balanced regime produces more coherent and structurally consistent reconstructions, closely matching the HR reference. 
    In the low-resolution but high-SNR setting, structures appear over-smoothed, and fine anatomical details are lost due to partial volume effects. 
    }
    \label{fig:snr_tradeoff}
\end{figure}

\IEEEpubidadjcol

Despite strong performance, most MRI super‑resolution methods~\cite{chaudhari2018super,chen2018brain,lyu2020multi} treat the task as a deterministic mapping from a given low‑resolution scan to a fixed high‑resolution target. This perspective assumes that the acquired low‑resolution image is an optimal and static starting point determined by the MRI system. In practice, however, each MRI scan represents only one of many possible realizations, and MRI image quality can vary substantially in both SNR and resolution. As dictated by MRI acquisition physics~\cite{plewes2012physics,currie2013understanding}, under fixed hardware and scan‑time constraints, spatial resolution and SNR are fundamentally intertwined: increasing resolution reduces SNR, while improving SNR can come at the cost of spatial resolution.
Furthermore, recent studies show that tuning acquisition parameters to achieve an SNR of around 16 dB captures the most structurally informative content~\cite{portnoy2009information}, especially for ultra‑low‑field MRI systems, as illustrated in Figure~\ref{fig:snr_tradeoff}. This underscores that the given low‑resolution image may be not the optimal one acquired from the MRI system.

These observations motivate a rethinking of MRI super-resolution. Rather than treating super-resolution as the post hoc upscaling of a fixed low-resolution input, we ask \textit{\textbf{whether MRI super-resolution can be formulated as a physics-aware reconstruction problem that is explicitly tied to the underlying imaging system.} }Under this paradigm, super-resolution should account for the acquisition process itself, where resolution, SNR, and sampling efficiency are inherently coupled and dynamically traded off. The goal is thus not merely to increase spatial resolution, but to reconstruct images with improved structural clarity and physical fidelity under given hardware and acquisition constraints.
However, realizing this paradigm is challenging: models must be capable of handling diverse acquisition regimes while preserving physically plausible and physiologically meaningful structures. 
Yet most existing MRI super-resolution methods~\cite{li2024rethinking_diffusion,li2022deep,zhao2020smore} are built for predefined, integer upsampling ratios, limiting their applicability to resolution‑heterogeneous acquisitions.

To address this limitation, we propose adapting the 2D Gaussian Splatting (2D GS) framework~\cite{peng2025pixel} to the MRI domain by treating the image as a continuous signal rather than a discrete pixel array. The core idea for supporting dynamic input resolution is to reformulate reconstruction as a coordinate‑based rendering problem: instead of operating on a fixed-size grid, 2D GS learns to map low-resolution inputs of arbitrary dimensions into a unified, continuous Gaussian field, thereby enabling flexible, resolution‑agnostic reconstruction.
However, directly applying 2D GS to MRI remains non-trivial due to three reasons: (1) \textit{\textbf{Lack of domain-specific priors:}} Unlike natural images, MRI data adheres to specific anatomical structures and acquisition-dependent ``fingerprints.'' Ignoring these priors typically leads to suboptimal structural fidelity and poor tissue definition.
(2) \textit{\textbf{Lack of biophysical plausibility:}} MRI signals are determined by biophysical properties, such as proton density and relaxation parameters ($T_1, T_2$), rather than independent RGB channels. Without enforcing biophysical constraints, reconstructions often lack physical validity and consistency.
(3) \textit{\textbf{{Limited} Data}}: Paired low- and high-resolution MRI datasets across diverse resolution settings are limited in practice, restricting the training of robust, resolution-agnostic models.

To address the first challenge, we introduce the \textit{\textbf{prior-aware Gaussian representation}}, which integrates anatomical and system-specific priors into the geometric modeling for MRI super-resolution. First, based on the observation that geometric complexity varies across different brain regions, we propose an Anatomical Structure Prior that initializes tissue-specific Gaussian kernel densities, ensuring that representational capacity is concentrated in anatomically intricate regions, such as cortical folds. Second, recognizing that MRI systems exhibit inherent global patterns (e.g., point spread functions and acquisition artifacts), we introduce an Imaging System Prior that captures these system-wide characteristics through a covariance dictionary. Specifically, this covariance dictionary models the discrepancy between raw MRI signals and the underlying anatomy, ensuring that the geometric shapes of Gaussian primitives remain consistent with the unique acquisition properties of the imaging hardware.

To address the biophysical consistency concern, we additionally introduce \textit{\textbf{physics-constrained signal modeling}} grounded in MRI acquisition physics. Rather than directly regressing pixel intensities, our framework predicts intrinsic tissue parameters—proton density ($\rho$) and effective relaxation rate ($R_2$)—and subsequently computes signal intensity through the governing physical equations. This formulation ensures that reconstructed images maintain biophysically plausible contrast relationships (e.g., CSF appearing hyperintense relative to gray matter in T2-weighted imaging). Furthermore, to overcome data scarcity, we develop a \textit{\textbf{meta-learning framework}} that enables effective utilization of rare paired low- and high-resolution real data by first pretraining the model on simulated data and then adapting it to real-world conditions through meta-learning.

The main contributions of this work are:
\begin{enumerate}
\item \textbf{Physics-aware MRI super-resolution framework.} We reformulate MRI super-resolution as a physics-aware reconstruction problem that explicitly models the resolution-SNR trade-off, enabling enhanced detail recovery beyond conventional fixed-scale approaches.

\item \textbf{2D Gaussian Splatting-based MRI super-resolution.} We pioneer the adaptation of 2D Gaussian Splatting to MRI super-resolution through multiple carefully designed modules, including segmentation-guided primitive initialization, an MRI-specific covariance dictionary, and physics-constrained signal modeling that ensures biophysically plausible reconstructions.

\item \textbf{State-of-the-art performance.} Our framework significantly outperforms all baseline methods on both dynamic-resolution datasets and the FastMRI benchmark in terms of quantitative metrics and qualitative comparisons, demonstrating considerable potential for practical clinical applications.
\end{enumerate}
\section{Related Work}
\subsection{MRI Super Resolution}
High-resolution magnetic resonance imaging (MRI) is crucial for accurate anatomical assessment, but its acquisition is constrained by the inherent trade-off among spatial resolution, signal-to-noise ratio (SNR), and scan time. Specifically, achieving higher spatial resolution typically requires longer scan times to maintain sufficient SNR. Low-resolution (LR) imaging is faster and may boost SNR, but often lacks the detail needed for high-quality diagnostics~\cite{khateri2025mri}. To alleviate the inherent trader-off, MRI super-resolution aims to reconstruct HR images from more accessible LR scans. MRI SR methods can be categorized into model-based approaches and learning-based approaches. Model-based methods explicitly formulate the image degradation process and recover the HR image by solving an inverse optimization problem. Within this framework, different approaches are mainly distinguished by the choice of handcrafted priors. Classical approaches include Tikhonov~\cite{zhang2008tikhonov,brudfors2019tool}, total variation(TV)~\cite{rudin1992nonlinear,tourbier2015efficient,shi2015lrtv}, self-similarity~\cite{glasner2009super,manjon2010mri,bustin2018isotropic}, low-rankness~\cite{shi2015lrtv,cherukuri2019deep,li2022motion,zhang2022accelerated}, sparse representation~\cite{zhang2015structure,wang2014sparse,zhang2012hierarchical}, non-local mean~\cite{protter2008generalizing,manjon2010nonlocal,jafari2014mri}, and gradient guidance~\cite{farsiu2004fast,sui2019isotropic,sui2020learning}. However, their effectiveness strongly depends on the assumed degradation model and the manually designed regularization terms, making them less robust to complex real-world degradations, especially in low-field MRI. To overcome these limitations, learning-based methods have emerged as a powerful alternative, which learn the mapping from LR to HR images directly from data without relying on explicit degradation models or handcrafted prior terms.
Recent work leverages generative model such as generative adversarial networks (GANs)~\cite{ledig2017srgan,wang2020enhanced,zhang2022soupgan,wang2023disgan,zhang2025srgan}, diffusion models~\cite{wang2023inversesr,safari2025mri}, Implicit Neural Representations (INRs)~\cite{xu2023nesvor,fang2024cycleinr}  and 2D Gaussian Splatting (2D GS)~\cite{hu2025gaussiansr,peng2025pixel} to enhance perceptual quality. 

Despite significant progress, existing MRI super-resolution methods predominantly frame the task as a deterministic mapping from a specific low-resolution input to a fixed high-resolution target, without considering the MRI acquisition system itself. In contrast, we propose a physics-aware MRI super-resolution framework that explicitly models the resolution-SNR trade-off, enabling enhanced detail recovery beyond conventional fixed-scale approaches.

\subsection{Continuous Super-Resolution}
Traditional super-resolution methods are typically designed for discrete and predefined integer upscaling factors, such as ×2, ×4, or ×8. While these approaches have achieved remarkable performance, they lack flexibility in handling continuous resolutions required in real-world scenarios. Additionally, separate models are often trained for different scaling factors, which significantly increases the computational cost. To address these limitations, recent works have explored arbitrary-scale super-resolution (ASSR) methods, which aim to train a unified model capable of handling arbitrary upscaling factors. For example, MetaSR~\cite{hu2019metasr} introduces a meta-learning-based framework to predict upsampling filters conditioned on arbitrary scale factors, enabling flexible resolution enhancement. LIIF~\cite{chen2021liif} models images as continuous functions by learning local implicit representations, allowing pixel-wise prediction at arbitrary coordinates. LTE~\cite{xu2022lte} further improves LIIF by incorporating local texture estimation to better capture high-frequency details. CiaoSR~\cite{cao2023ciaosr} enhances coordinate-based SR with improved feature aggregation strategies, achieving more accurate continuous reconstruction. However, these implicit modeling methods struggle to explicitly capture continuous signal structures and rely on time-consuming upsampling and decoding processes, resulting in suboptimal efficiency and limited generalization capability ~\cite{peng2025pixel}. Recently, 2D Gaussian Splatting (2D GS) ~\cite{hu2025gaussiansr,peng2025pixel} has emerged as a promising alternative for continuous super-resolution, demonstrating superior performance over previous methods. Motivated by this, we adapt 2D GS to MRI super-resolution by introducing prior-aware Gaussian representation and a physically-constrained intensity module, enabling better structural clarity and contrast fidelity.

\section{Preliminary}
\label{sec:preliminary}
\subsection{Resolution-SNR Trade-off in MRI}
\label{sec:motivation}


We first clarify the resolution-SNR trade-off in MRI, which forms the foundation of our work. The MRI signal originates from hydrogen protons in the body, and the detected signal from each voxel represents the sum of contributions from all protons within it. Thus, signal amplitude is approximately proportional to voxel volume. When voxel size is reduced to increase spatial resolution, signal amplitude decreases proportionally. However, noise arises primarily from the receiver coils and remains constant regardless of voxel size. Because noise does not scale down with voxel volume as signal does, smaller voxels exhibit lower SNR, creating an intrinsic trade-off between spatial resolution and SNR.

Formally, we start from the basic MRI signal equation to derive this resolution--SNR trade-off. Specifically, taken the most clinical-widely used sequence Spin Echo sequence~\cite{jung2013spin} as example, the measured MRI signal of a single hydrogen spin ($S_{hp}$) can be given by:
\begin{equation}
S_{hp} = C \cdot B_0 \cdot 
\left( 1 - e^{-\dfrac{TR}{T_1(B_0)}} \right)
\cdot 
e^{-\dfrac{TE}{T_2(B_0)}},
\end{equation}
where $C$ is a system-dependent constant incorporating hardware sensitivity and physical constants, $B_0$ is the main magnetic field strength, $TR$ is the repetition time, $TE$ is the echo time, and $T_1(B_0)$ and $T_2(B_0)$ are the longitudinal and transverse relaxation times, respectively.

For a voxel of volume $V_\mathrm{voxel} = \Delta x \cdot \Delta y \cdot \Delta z$, the total number of contributing spins scales linearly with $V_\mathrm{voxel}$, so the signal amplitude is given by:
\begin{equation}
    S_{\mathrm{voxel}}
    \propto
    \int_{V_{\mathrm{voxel}}}
    \rho(\mathbf{r})\,S_{hp}\,d\mathbf{r},
\label{eq:voxel_signal}
\end{equation}
where $\rho(\mathbf{r})$ is the proton density (the number of hydrogen spins per unit volume) at position $r$.
Consequently, when $\Delta x,\, \Delta y,\, \Delta z$ are small, $S_{\mathrm{voxel}}$ scales approximately linearly with the voxel volume.

Meanwhile, the noise introduced by system hardware during acquisition remains approximately independent of voxel size and decreases only with signal averaging and acquisition time:
\begin{equation}
    \sigma_n
    \propto
    \frac{1}{\sqrt{N_{ex}T_{acq}}},
\label{eq:noise_scaling}
\end{equation}
where $N_{ex}$ is the number of scanning repetitions and $T_{acq}$ is acquision time.

From Eqs.~(\ref{eq:voxel_signal})--(\ref{eq:noise_scaling}), 
the classical MRI SNR expression becomes~\cite{brown2014magnetic}:
\begin{equation}
\mathrm{SNR}
\propto \frac{S_{\mathrm{voxel}}}{\sigma_n}
\propto \frac{\bar{\rho}\Delta x\Delta y\Delta z}{\displaystyle\frac{1}{\sqrt{N_{ex}T_{acq}}}}
= \bar{\rho}\Delta x\Delta y\Delta z\sqrt{N_{ex}T_{acq}},
\label{eq:snr_full}
\end{equation}
where $\bar{\rho}$ is the average proton density within the small voxel. {This Eq.~\ref{eq:snr_full} directly reveals the trade-off relationship between resolution and SNR, which significantly impact the MRI image quality as shown in Fig.~\ref{fig:snr_tradeoff}. This motivates us to search optimal resolution image for MRI super-resolution.} For more details, see in Section S1 of Supplementary Materials.

\subsection{2D Gaussian Splatting for Super-Resolution}

Achieving such dynamic super-resolution requires an image representation that is inherently continuous and resolution-agnostic. 2D Gaussian Splatting (2D GS)~\cite{peng2025pixel} offers precisely this capability by representing images as collections of geometric primitives rather than discrete pixel grids. We adopt 2D GS as the foundation of our method and briefly review its formulation here.

The core idea of 2D GS is to represent an image using $N$ anisotropic Gaussian primitives, each defined by learnable spatial and appearance parameters. This continuous representation naturally supports arbitrary-scale super-resolution without retraining. The framework operates in three stages: (1) encoding a low-resolution input into continuous feature representations, (2) predicting the properties of Gaussian primitives from these features, and (3) rendering the high-resolution output image. We detail each stage below.


\noindent\textbf{Feature Extraction.}
Given a low-resolution (LR) input image $I_{LR}$, a backbone encoder $E(\cdot)$ is first employed to extract deep feature representations:
\begin{equation}
F = E(I_{LR}),
\end{equation}
These features encode both local texture details and global structural information, providing the foundation for subsequent Gaussian parameter prediction.

\noindent\textbf{Gaussian Parameter Prediction.}
The parameters of all Gaussian primitives are predicted from the feature representation $F$ via a set of lightweight multilayer perceptrons (MLPs):
\begin{equation}
\{\boldsymbol{\mu}_i, \boldsymbol{\Sigma}_i, \mathbf{c}_i\}_{i=1}^{N} = \text{MLP}(F),
\end{equation}
where $\boldsymbol{\mu}_i \in \mathbb{R}^2$ denotes the spatial position, $\mathbf{c}_i \in \mathbb{R}^3$ denotes the color, and $\boldsymbol{\Sigma}_i \in \mathbb{R}^{2 \times 2}$ denotes the covariance matrix of the $i$-th Gaussian primitive. Specifically, the color parameter represents the amplitude of the Gaussian distribution, corresponding to pixel intensity values (e.g., three channels for RGB images). The covariance matrix controls the spatial extent and orientation of the Gaussian kernel, enabling anisotropic modeling of local image structures.


\noindent\textbf{Image Rendering.}
Once the Gaussian primitives are defined, a high-resolution image is rendered by these primitives:
\begin{equation}
I(\mathbf{x}) = \sum_{i=1}^{N} \mathcal{G}(\mathbf{x}; \boldsymbol{\mu}_i, \boldsymbol{\Sigma}_i, \mathbf{c}_i),
\end{equation}
High-resolution images at arbitrary scales can be obtained by discretizing this field onto a target coordinate grid.

\noindent\textbf{Training Objective.}
The rendering process is fully differentiable, allowing the entire pipeline to be optimized end-to-end. Given paired LR--HR training data, the model is supervised using a reconstruction loss:
\begin{equation}
\mathcal{L}_{rec} = \| I(x) - I_{HR} \|_1.
\end{equation}

\noindent\textbf{Limitations for MRI Application.} While 2D GS achieves strong performance on natural images, direct application to MRI presents several challenges:
\begin{itemize}
\item \textbf{Lack of domain-specific priors}: Position and covariance estimation without guidance from anatomical and imaging system priors may generate geometrically unreliable primitives, undermining structural coherence.
\item \textbf{Lack of biophysical plausibility}: Direct intensity prediction without MRI-physics contrains may produce physically implausible tissue contrasts, compromising downstream analysis quality.
\item \textbf{Limited data}: Paired low- and high-resolution MRI images are scarce. Although this is not a limitation specific to 2D GS, this MRI-specific challenge still needs to be addressed, as it poses significant difficulties for effective model optimization.
\end{itemize}

\section{Method}
\label{sec:method}


As discussed in the preliminary section, our goal is to achieve dynamic adjustment along the resolution-SNR spectrum, enabling MRI super-resolution at informatively optimal operating points. Moreover, while 2D Gaussian Splatting (2D GS) provides a suitable framework that inherently supports continuous input resolution, the limitations identified above hinder the direct application of vanilla 2D GS to this task.

\begin{figure*}[t]  
    \centering
    \includegraphics[width=\textwidth]{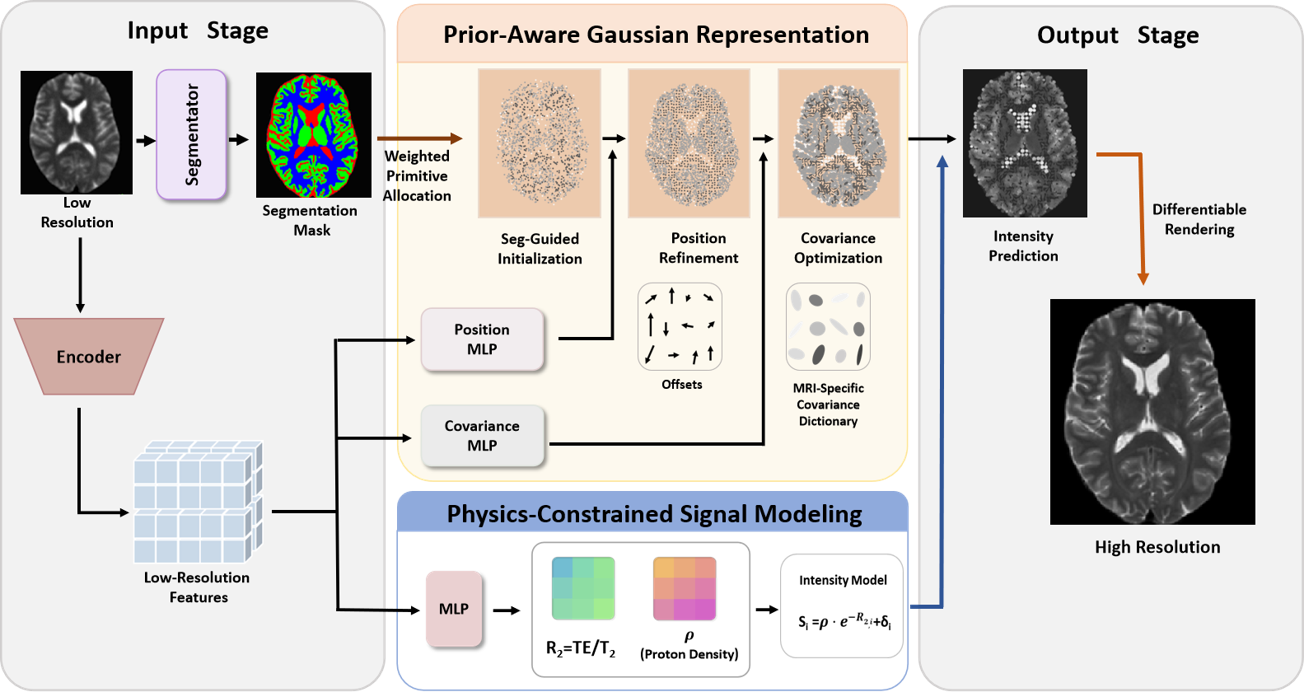} 
    \caption{
        Overview of the proposed physics-aware 2D GS framework. An arbitrary-resolution input is processed through two parallel pathways: a segmentator generates tissue masks for segmentation-guided primitive initialization, while an encoder extracts low-resolution features. Two modules subsequently estimate Gaussian parameters: \textit{prior-aware representation} predicts position offsets and selects covariance matrices from an MRI-specific dictionary, while \textit{physics-constrained signal modeling} computes intensities from tissue properties ($\rho$, $R_2$) via the MRI signal equation. The final high-resolution output is rendered through differentiable splatting.
    }
    \label{fig:pipeline_draft_2}
\end{figure*}

To address these challenges, we propose a physics-aware Gaussian splatting framework that enables continuous-scale MRI super-resolution. Specifically, as illustrated in Figure~\ref{fig:pipeline_draft_2}, our approach introduces three targeted innovations: (1) \textbf{Prior-aware Gaussian representation} (Section~\ref{sec:prior_aware}) that insert domian-specific priors to guarantee anatomy structural fidelity and clarity, (2) \textbf{Physics-constrained signal modeling} (Section~\ref{sec:physics}) that designs a tissue relaxation parameters-based signal equation for super-resolution to ensures biophysically plausible, and (3) \textbf{Meta-learning-based adaptation} (Section~\ref{sec:meta}) that bridges the domain gap between synthetic training and real low-field acquisitions.

\subsection{Prior-Aware Gaussian Representation}
\label{sec:prior_aware}

Effective Gaussian splatting relies on appropriate primitive initialization and covariance parameterization. However, the original 2D GS determines these spatial positions and geometric properties without domain-specific guidance, employing grid-based uniform initialization and computing covariance parameters according to natural image characteristics, leading to suboptimal performance. Here, we introduce two critical priors in MRI: anatomical structure and imaging system fingerprint. The anatomical structure prior models structural complexities of different tissues, assisting model to easier initialize and refine the Gaussian primitive's position.
And the imaging system fingerprint refers to the global MRI system pattern, such as point spread function and acquisition artifacts, helping model to effectively optimize the final covariance of  Gaussian primitives. We details how to utilize each prior below.

\subsubsection{Anatomical structure guided initialization and refinement}
\label{sec:seg_init}
From anatomy perspective, different brain tissues are assumed to have different structural complexities, e.g. in brain, the structure of components, gray matter (GM), white matter (WM) is relatively complex and cerebrospinal fluid (CSF) tends to be simpler.
Given this, we propose a segmentation-guided initialization that places more primitives to structurally complex regions, ensuring adequate representational capacity where it is most needed.

Specifically, a segmentator is applied to generate a tissue segmentation mask that partitions the image into $K$ regions $\{\Omega_k\}_{k=1}^{K}$ (e.g., CSF, gray matter, white matter, and background for brain). For each region, we assign a density weight that jointly accounts for two factors: the region's spatial extent and its structural complexity. Intuitively, tissues with more intricate structures should be represented with a denser set of Gaussian primitives, while larger regions should also receive proportionally more primitives to adequately cover their area. 

Concretely, the density weight for region $\Omega_k$ is defined as
\begin{equation}
d_k = \frac{a_k \cdot w_k}{\sum_{j=1}^{K} a_j \cdot w_j},
\end{equation}
where $a_k = \frac{|\Omega_k|}{\sum_j |\Omega_j|}$ is the normalized area ratio, and $w_k$ is a predefined complexity weight reflecting the morphological intricacy of each tissue type. We set $w_{\text{GM}} > w_{\text{WM}} > w_{\text{CSF}}$, based on the observation that gray matter usually exhibits the most complex folding patterns, white matter has moderate complexity, and CSF regions are relatively homogeneous. Given a total budget of $N$ primitives, region $\Omega_k$ receives
\begin{equation}
N_k = \big\lfloor d_k \cdot N \big\rfloor,
\end{equation}
primitives. This design ensures that the Gaussian kernel sampling density is adaptively higher for tissues that are both more complex and occupy a larger area, leading to a more faithful and efficient representation of different brain structures.

Although initialization determines anatomically meaningful starting positions, optimal primitive placement may require further adjustment to capture local image content. Thus, we propose a position refinement module $\mathcal{M}_{\mathrm{pos}}$ that predicts per-primitive offsets conditioned on the extracted feature map:
\begin{equation}
    \boldsymbol{\mu}_i
    = \boldsymbol{\mu}_i^{(0)} 
    + \delta \,\tanh\big(\mathcal{M}_{\mathrm{pos}}(F_{\mathrm{LR}})\big)_i,
\end{equation}
where $\tanh$ constrains the normalized offsets to $[-1, 1]$, and the scaling factor $\delta$ controls the maximum displacement range $[-\delta,\delta]$, preventing excessive drift from the anatomically grounded initial positions. This two-stage design combines anatomical structure priors with data-driven refinement, allowing primitives to adapt to patient-specific variations while maintaining anatomical plausibility.

\subsubsection{Image system-prior based Covariance optimization}

As shown in Figure~\ref{fig:pipeline_draft_2}, following position refinement, we proceed to optimize the kernel geometry properties. Our key insight is that when a Gaussian primitive is optimized to represent a local region in an MRI image, its geometry reflects not only the underlying anatomical structure but also the imaging characteristics introduced by the MRI acquisition system. Consequently, the covariance matrices are inherently influenced by imaging system priors.

Motivated by this observation, we introduce an MRI-Specific Covariance Dictionary that records the covariance distribution of high-resolution MRI images. Specifically, combined with the anatomical structure designs introduced above, this dictionary effectively captures MRI system patterns (such as acquisition artifacts and point spread functions), ensuring that our model generates structures consistent with hardware-specific characteristics.

To build our dictionary, we collected a large number of high-resolution brain MRI slices and fitted each with 2D GS using standard optimization. From these fitted primitives, we extracted the covariance parameters $(\sigma_x^2, \sigma_y^2, \rho\sigma_x\sigma_y)$ and analyzed their empirical distributions. The parameters exhibit characteristic that are notably distinct from those observed in natural images, reflecting the unique spatial statistics imposed by MRI physics. We then constructed a dictionary of $M$ representative covariance matrices $\{\boldsymbol{\Sigma}_j^{\text{dict}}\}_{j=1}^{M}$ by sampling from these empirical distributions.

During super-resolution, rather than directly predicting covariance entries, the covariance module $\mathcal{M}_{\mathrm{cov}}$ outputs combination weights over dictionary elements:
\begin{equation}
\boldsymbol{\Sigma}_i = \sum_{j=1}^{M} \text{Softmax}(\mathcal{M}_{\mathrm{cov}}(F_{\mathrm{LR}}))_{i,j} \cdot \boldsymbol{\Sigma}_j^{\text{dict}}.
\end{equation}

This formulation constrains the covariance space to MRI-realistic configurations, improving training stability while preserving flexibility for diverse anatomical structures. 

\subsection{Physics-Constrained Signal Modeling}
\label{sec:physics}

The prior-aware representation introduced above determines \textit{where} Gaussian primitives are placed and \textit{how} they are shaped. Now we turn to the question of \textit{what intensity values} these primitives should carry. 

In MRI, image intensity is not an arbitrary quantity but is governed by well-defined biophysical processes described by the Bloch equations. Therefore, directly predicting pixel intensities neglects this underlying physics and may produce outputs that violate expected tissue contrast relationships. To address this, we introduce physics-constrained signal modeling that grounds intensity prediction in tissue relaxation properties and local proton density, ensuring that reconstructed images remain consistent with MRI signal formation principles.

\subsubsection{Signal Formation Model}

MRI signal intensity is fundamentally determined by tissue-specific relaxation properties and acquisition parameters. For T2-weighted imaging, the signal can be modeled as:
\begin{equation}
S(\rho, T_1, T_2) = \rho \cdot (1 - e^{-TR/T_1}) \cdot e^{-TE/T_2},
\end{equation}
where $\rho$ is proton density, $T_1$ and $T_2$ are longitudinal and transverse relaxation times, and $TR$, $TE$ are the repetition and echo time, respectively. In typical T2-weighted acquisitions, $TR$ is chosen to be sufficiently long relative to $T_1$, such that the $T_1$-dependent term approaches unity. Under this condition, the signal simplifies to:
\begin{equation}
S \approx \rho \cdot e^{-TE/T_2} = \rho \cdot e^{-R_2},
\end{equation}
where $R_2 = TE/T_2$ represents the effective relaxation term. This equation captures the dominant contrast mechanism in T2-weighted images and serves as the physical basis for our intensity prediction.

\subsubsection{Tissue Parameter Prediction}

Rather than directly regressing pixel intensities, we leverage the signal model above to decompose intensity prediction into physically meaningful components. Specifically, we interpret each Gaussian primitive as representing a localized proton spin ensemble and predict its intrinsic tissue parameters:
\begin{equation}
[\rho_i, R_{2,i}] = \mathcal{M}_{\mathrm{signal}}(F_{\mathrm{LR}}),
\end{equation}
where $\rho_i$ and $R_{2,i}$ denote the proton density and effective relaxation term for primitive $i$. The final primitive intensity is then computed through a physics-constrained intensity layer:
\begin{equation}
c_i = \rho_i \cdot e^{-R_{2,i}} + \delta_i,
\end{equation}
where $\delta_i$ is a learnable residual that accommodates modeling approximations and acquisition imperfections not captured by the idealized signal equation.

This physics-grounded formulation ensures that predicted intensities arise from tissue-specific parameters rather than unconstrained regression, thereby maintaining biophysically plausible contrast relationships. Moreover, since $\rho$ and $R_2$ reflect intrinsic tissue properties that are independent of imaging resolution, this decomposition encourages the network to learn resolution-invariant representations, improving generalization across different super-resolution scales. 

\subsection{Model Training and Meta-Learning for Domain Adaptation}
\label{sec:meta}


Building upon the MRI-specific network architecture introduced previously, we now describe the model training and meta-learning pipeline.

Given a large simulated low-field MRI dataset $\mathcal{D}_{sim}$ (e.g., comprising multiple resolution-SNR configurations of 64mT images) and a limited amount of paired real low-field MRI data $\mathcal{D}_{real}$, our training pipeline operates as follows. We first employ a segmentator to predict the segmentation mask $\mathcal{D}_{mask}$ from the low-resolution input of $\mathcal{D}_{sim}$. With the synthetic paired data $\mathcal{D}_{sim}$ and corresponding anatomical priors $\mathcal{D}_{mask}$, a conventional two-stage training paradigm (i.e., pre-training our framework on simulated data followed by fine-tuning on $\mathcal{D}_{real}$) would encounter two critical limitations. First, models pre-trained exclusively on simulated degradations often generalize poorly to real low-field acquisitions, as the complex imaging characteristics of ultra-low-field MRI are inherently difficult to simulate realistically. Second, when encountering novel degradation conditions or imaging protocols, such models require extensive fine-tuning from scratch, which is time-consuming and impractical for clinical deployment.

To address these limitations, we propose a meta-learning extension that replaces the standard training paradigm with episodic meta-training, jointly leveraging both $\mathcal{D}_{sim}$ and $\mathcal{D}_{real}$. Inspired by MAML~\cite{nichol2018first}, we reformulate the training objective: rather than learning a fixed low-to-high-field mapping, we learn a shared initialization that can rapidly adapt to unseen low-field conditions with only a few gradient steps. This meta-learning framework enables rapid adaptation to novel degradation conditions, even in previously unseen settings, significantly reducing fine-tuning costs while improving generalization to real low-field MRI acquisitions, as demonstrated in Section. V-C of Supplementary Materials.

\section{Experiments}
\subsection{Implementation Details}
\label{sec:imp_details}
\textbf{Datasets.}
\label{sec:dataset}
To investigate our proposed dynamic-resolution hypothesis, we use T2-weighted 3T MRI scans from the \textbf{IXI dataset}~\cite{ixi_dataset} as Ground Truth and simulate them to generate low-resolution (LR) data corresponding to a 64mT field. Specifically, the simulated LR data include multiple resolution levels (scales of $1.0$, $0.9$, $0.7$, $0.5$, and $0.25$), aiming to mimic real 64mT field MRI scans acquired with different spatial resolutions. Details of the simulation procedure and degradation parameters are provided in Section S2.11 of the Supplementary Material.
Following the experimental protocol established in ~\cite{wang2025diffusion}, we use the \textbf{fastMRI dataset} as the benchmark, selecting 960 T2-weighted axial slices for training and 145 slices for testing. In addition, low-resolution (LR) inputs are simulated from the selected slices via k-space truncation.
Specifically, for a given scaling factor $s \in \{4, 5, 6.4\}$, only the central $1/s$ portion of k-space is retained. all degraded images are resampled to $320 \times 320$ using bicubic interpolation.
To demonstrate that our framework generalizes to authentic ultra-low-field data, we adopt a publicly available \textbf{paired 64mT--3T dataset} released by Leiden University Medical Center~\cite{van_den_broek_2024_15862148} for training and testing. This dataset contains paired low-field (64mT) and high-field (3T) brain MRI scans from 11 healthy subjects. We include 210 axial T2 slices for meta-learning-based domain adaptation experiments, as detailed in Section S3.3 of the Supplementary material.

\textbf{Training setup.}
Following previous work~\cite{peng2025pixel}, we employ SwinIR ~\cite{liang2021swinir} as the encoder backbone and 2D U-Net ~\cite{ronneberger2015u} as the segmentator for all experiments. Adam ~\cite{kingma2014adam} is used as the optimizer, with the initial learning rate $1 \times 10^{-4}$ and decaying by a factor of 0.5 every 100 epochs. We utilize  a combination of L1 loss, gradient loss and frequency loss for training, with a total batch size of 4 and 1,000 epochs on 1 NVIDIA RTX 4090 GPU.

\subsection{Experimental Setup}
\label{sec:exper_setup}

In this paper, we set up two types of experiments, focusing on \textbf{dynamic-resolution} and \textbf{static-resolution} settings. To investigate our proposed dynamic-resolution hypothesis, we design experiments on two datasets: a simulated 64 mT–3T paired dataset that we construct, and a real multi-resolution dataset consisting of paired 3T and 5T MRI scans. Furthermore, we also conduct experiments to evaluate the performance of our model under the static-resolution condition using two datasets: a simulated dataset derived from fastMRI and a real paired 64 mT–3 T dataset.

\begin{table}[t]
\centering
\caption{Quantitative results on simulated and real multi-resolution datasets.}
\label{tab:quantitative_results}
\footnotesize
\setlength{\tabcolsep}{3pt}
\renewcommand{\arraystretch}{0.92}
\resizebox{\columnwidth}{!}{
\begin{tabular}{lcccc}
\toprule
Method / Resolution 
& SSIM$\uparrow$ 
& PSNR$\uparrow$ 
& DISTS$\downarrow$ 
& HFEN$\downarrow$ \\
\midrule

\multicolumn{5}{c}{\textbf{Simulated IXI Dataset}} \\
\midrule
LIIF ($\times 1.0$)              & 0.8743 & 25.20 & 0.1968 & 0.9248 \\
LTE ($\times 1.0$)               & 0.8772 & 25.39 & 0.1969 & 0.6498 \\
Pixel-to-Gaussian ($\times 1.0$) & 0.8607 & 24.68 & 0.2175 & 0.7121 \\
\midrule
Ours ($\times 1.0$)              & 0.9021 & 28.06 & 0.1231 & 0.3157 \\
Ours ($\times 0.9$)              & 0.9179 & 27.98 & 0.1231 & 0.3097 \\
\rowcolor{gray!20}
\textbf{Ours ($\times 0.7$)}     & \textbf{0.9234} & \textbf{28.10} & \textbf{0.1148} & \textbf{0.3051} \\
Ours ($\times 0.5$)              & 0.9216 & 27.87 & 0.1180 & 0.3096 \\
Ours ($\times 0.25$)             & 0.8791 & 26.02 & 0.1599 & 0.4799 \\

\midrule
\multicolumn{5}{c}{\textbf{Real Multi-Resolution Dataset}} \\
\midrule
LIIF ($\times 1.04$)              & 0.8745 & 25.1987 & 0.1279 & 0.5549 \\
LTE ($\times 1.04$)               & 0.8590 & 25.9498 & 0.1291 & 0.4979  \\
Pixel-to-Gaussian ($\times 1.04$) & 0.8387 & 24.1312 & 0.1283 & 0.5644 \\
\midrule
Ours ($\times 1.04$)  & \textbf{0.8775} & \textbf{26.7837} & 0.1274 & 0.4647 \\
Ours ($\times 0.90$)  & 0.8752          & 26.6568          & 0.1236 & 0.4768 \\
Ours ($\times 0.83$)  & 0.8706          & 26.5472          & \textbf{0.1227} & 0.4836 \\
\rowcolor{gray!20}
\textbf{Ours ($\times 0.76$)}  & \textbf{0.8775} & 26.7648 & 0.1246 & \textbf{0.4570} \\
Ours ($\times 0.69$)  & 0.8650          & 26.5186          & 0.1276 & 0.4855 \\
Ours ($\times 0.625$) & 0.8471          & 25.6540          & 0.1405 & 0.5565 \\
Ours ($\times 0.55$)  & 0.8627          & 26.2193          & 0.1569 & 0.4870 \\
Ours ($\times 0.50$)  & 0.8480          & 25.8045          & 0.1721 & 0.5269 \\

\bottomrule
\end{tabular}
}
\end{table}

\begin{table}[t]
\centering
\caption{Quantitative comparison on the fastMRI brain benchmark. Best results are in \textbf{bold}.}
\label{tab:fastmri_comparison}
\small
\setlength{\tabcolsep}{4pt}
\renewcommand{\arraystretch}{1.1}
\begin{tabular}{l|cc|cc|cc}
\toprule
\multirow{2}{*}{Method} 
& \multicolumn{2}{c|}{$4\times$} 
& \multicolumn{2}{c|}{$5\times$} 
& \multicolumn{2}{c}{$6.4\times$} \\
\cmidrule{2-7}
& PSNR$\uparrow$ & SSIM$\uparrow$ 
& PSNR$\uparrow$ & SSIM$\uparrow$ 
& PSNR$\uparrow$ & SSIM$\uparrow$ \\
\midrule
SwinIR        & 31.47 & 0.924 & 29.41 & 0.885 & 27.21 & 0.867 \\
MetaSR        & 30.21 & 0.921 & 29.56 & 0.887 & 27.48 & 0.886 \\
ArSSR         & 31.56 & 0.922 & 30.04 & 0.900 & 28.05 & 0.890 \\
McMRSR        & 31.62 & 0.924 & 30.28 & 0.902 & 28.72 & 0.894 \\
ScASSR        & 31.84 & 0.925 & 30.56 & 0.904 & 28.65 & 0.895 \\
SA-INR        & 31.69 & 0.924 & 30.24 & 0.901 & 28.41 & 0.892 \\
Score-MRI     & 30.02 & 0.921 & 30.12 & 0.899 & 26.70 & 0.857 \\
R2D2+         & 30.65 & 0.924 & 30.59 & 0.898 & 27.29 & 0.865 \\
SR3           & 31.50 & 0.927 & 30.12 & 0.899 & 28.42 & 0.883 \\
DiracDiff     & 30.88 & 0.923 & 30.02 & 0.883 & 29.31 & 0.867 \\
DisC-Diff     & 31.99 & 0.928 & 30.75 & 0.904 & 28.90 & 0.896 \\
DiffMSR       & 32.15 & 0.932 & 30.92 & 0.906 & 29.15 & 0.897 \\
SS-PRDDiff    & 32.19 & 0.937 & 31.24 & 0.907 & 29.36 & 0.897 \\
MS-PRDDiff    & 32.81 & 0.942 & 31.90 & 0.914 & 29.80 & 0.905 \\
Pixel-to-Gau    & 33.62 & 0.960 & 31.42 & 0.904 & 29.09 & 0.899 \\
\midrule
\textbf{Ours} & \textbf{34.26} & \textbf{0.962} 
              & \textbf{32.20} & \textbf{0.946} 
              & \textbf{30.06} & \textbf{0.917} \\
\bottomrule
\end{tabular}
\end{table}

\textbf{Dynamic-Resolution Experiments}:
We design dynamic-resolution training and testing on two datasets. Following the prior work on arbitrary-scale super-resolution (ASSR), including MetaSR~\cite{hu2019metasr}, LTE~\cite{xu2022lte}, LIIF~\cite{chen2021liif}, CiaoSR~\cite{cao2023ciaosr}, we conduct an experiment using multi-resolution data from the simulated 64mT-3T paired dataset to evaluate our proposed 

\begin{figure*}[t]
\centering
\includegraphics[width=0.9\textwidth]{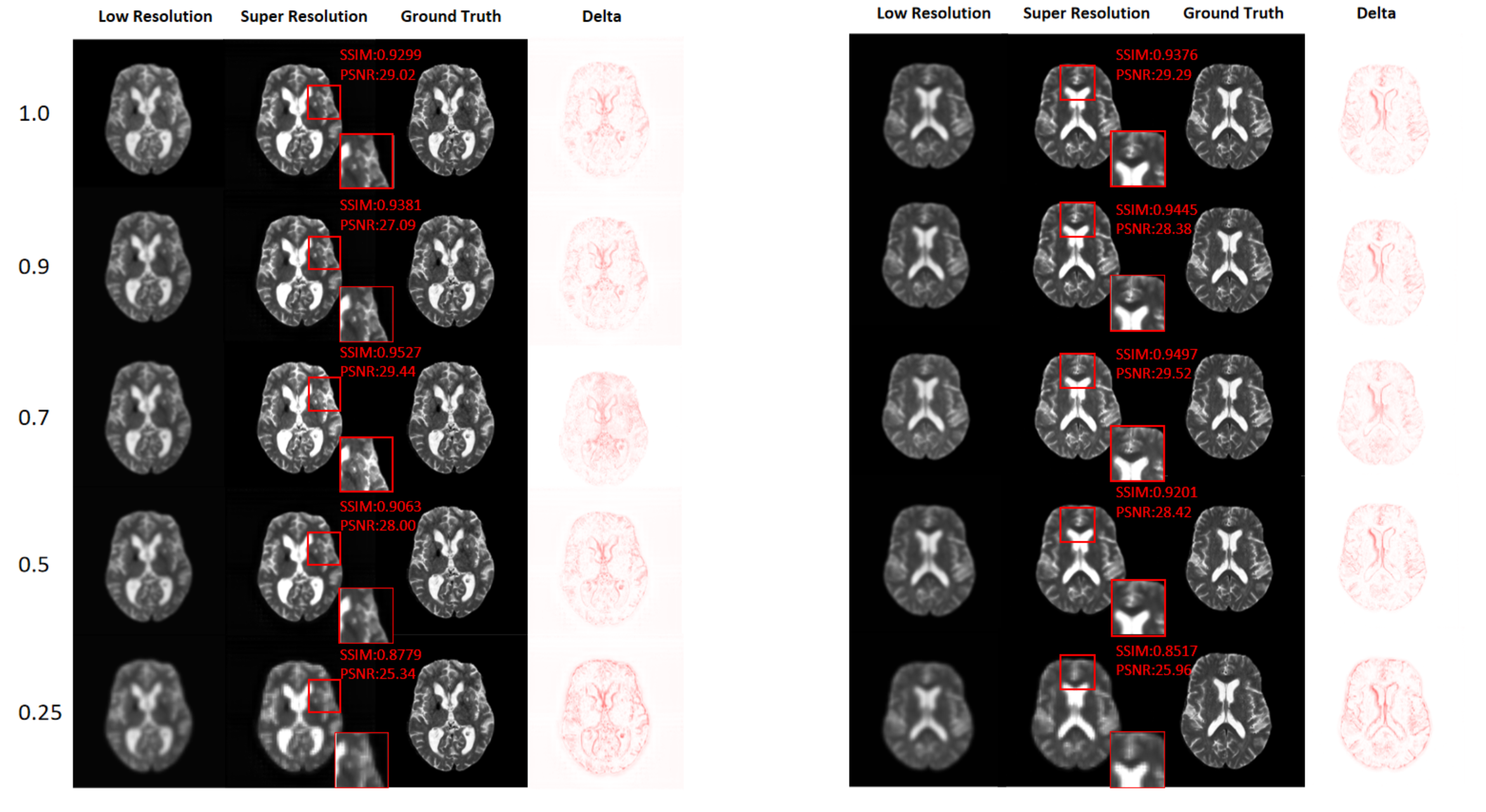}
\caption{Effect of input resolution on super-resolution quality on the simulated IXI dataset. Two representative cases are shown with varying scales from $\times 1.0$ to $\times 0.25$. The optimal visual quality is achieved at $\times 0.7$, exhibiting sharper edges and richer details. Red boxes highlight regions for detailed comparison.  }
\label{fig:fig_4}  
\end{figure*}

\begin{figure*}[t]
\centering
\includegraphics[width=0.9\textwidth]{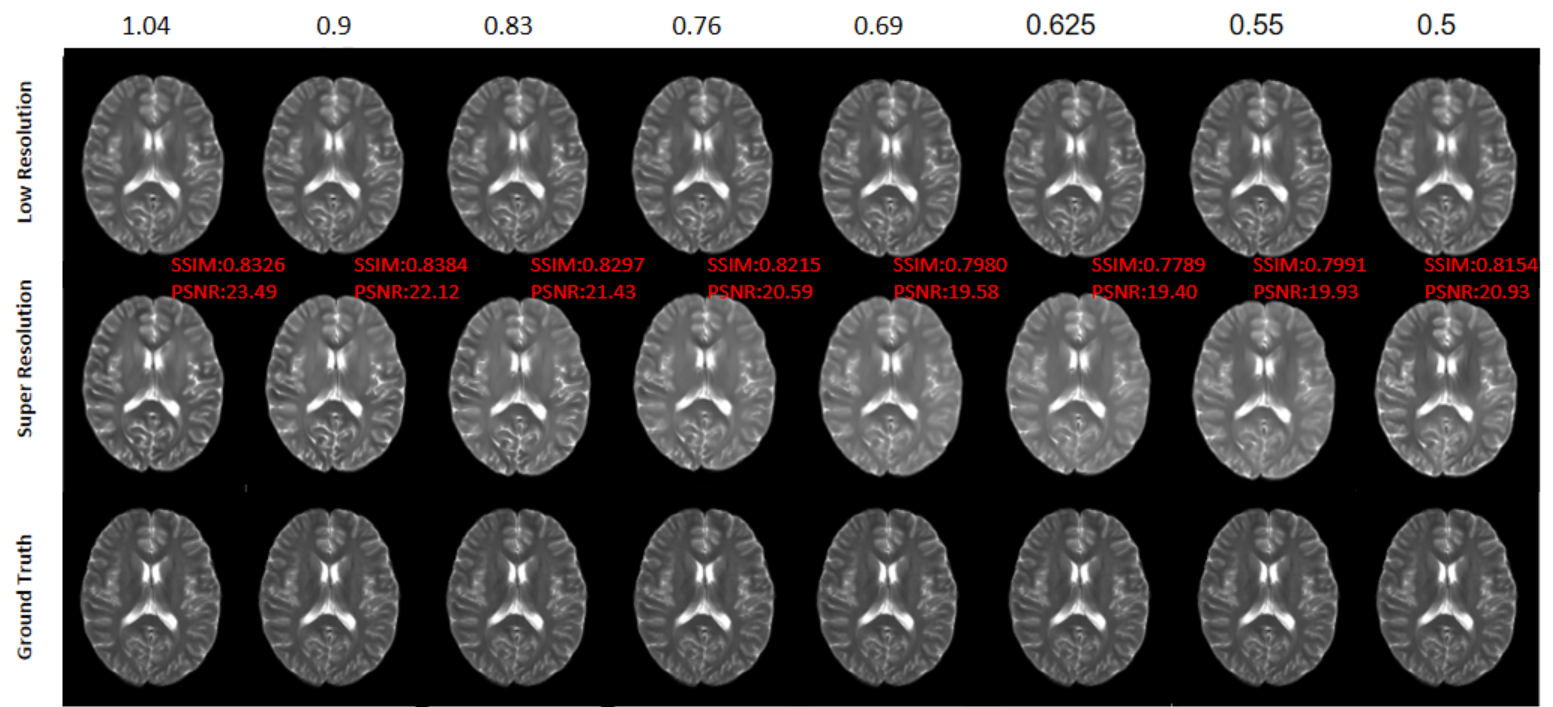}
\caption{Visual comparison on the real 3T--5T dynamic-resolution dataset. Our method produces reconstructions most consistent with the 5T Ground Truth, with improved structural details and realistic textures.}
\label{fig:3T5T}  
\end{figure*}
dynamic-resolution hypothesis. For each resolution scale, we randomly sample 1225 slice pairs for training, 350 pairs for validation, and 175 pairs for testing. Details of the slice selection procedure are described in the S2 of the Supplementary material. To validate our findings on real-world data, we conduct an experiment on a real-world dynamic-resolution dataset collecting from a 3T United Imaging uMR 890 device and a 5T United Imaging uMR Jupiter device. All data collection was approved by the Ethics Committee of ShanghaiTech Universitty (Approval No. Q2026-015), with further details provided in supplymentary S5. The dataset consists of multiple resolution 3T data (resolution scales of $1.04$, $0.9$, $0.83$, $0.76$, $0.69$, $0.625$, $0.55$, and $0.5$) and full-resolution 5T data as ground truth, collected from 10 healthy subjects. Details of the acquisition protocol and scanning parameters are provided in the S2 of the Supplementary material. We perform comparison with the following methods: LIIF\cite{chen2021liif}, LTE\cite{xu2022lte}, and Pixel-to-Gaussian\cite{peng2025pixel}. All comparison methods are trained using their officially released optimal experimental settings.

\textbf{Static-Resolution Experiments}:
We select simulated paired dataset as benchmark from fastMRI, comparing against 14 state-of-the-art methods: traditional SR methods (SwinIR~\cite{liang2021swinir}, MetaSR~\cite{hu2019metasr}), MRI-specific methods (ArSSR~\cite{chen2025dance}, McMRSR~\cite{li2022wavtrans}, ScASSR~\cite{li2022wavtrans}, SA-INR~\cite{wang2024implicit_mri}), and diffusion-based methods (Score-MRI~\cite{chung2022diffusion_mri}, R2D2+~\cite{chung2022diffusion_mri}, SR3~\cite{saharia2023sr3}, DiracDiff~\cite{fabian2024diracdiffusion}, DisC-Diff~\cite{mao2023discdiff}, DiffMSR~\cite{li2024rethinking_diffusion}, SS-PRDDiff~\cite{wang2025diffusion}, MS-PRDDiff~\cite{wang2025diffusion}). Following the benchmark protocol in~\cite{wang2025diffusion}.
Inspired by~\cite{he2024metags}, we employ meta-learning to alleviate the domain gap between simulated data and real data. The detailed task design for meta-learning and the meta-optimization strategy are provided in Section S3.3. In this experiment, we use 920 simulated paired data slices and 180 real paired data slices for training, and 60 real paired data slices for testing. We compare our method with LIIF\cite{chen2021liif}, LTE\cite{xu2022lte}, and Pixel-to-Gaussian\cite{peng2025pixel}. 

\begin{figure*}[t]
\centering
\includegraphics[width=\linewidth]{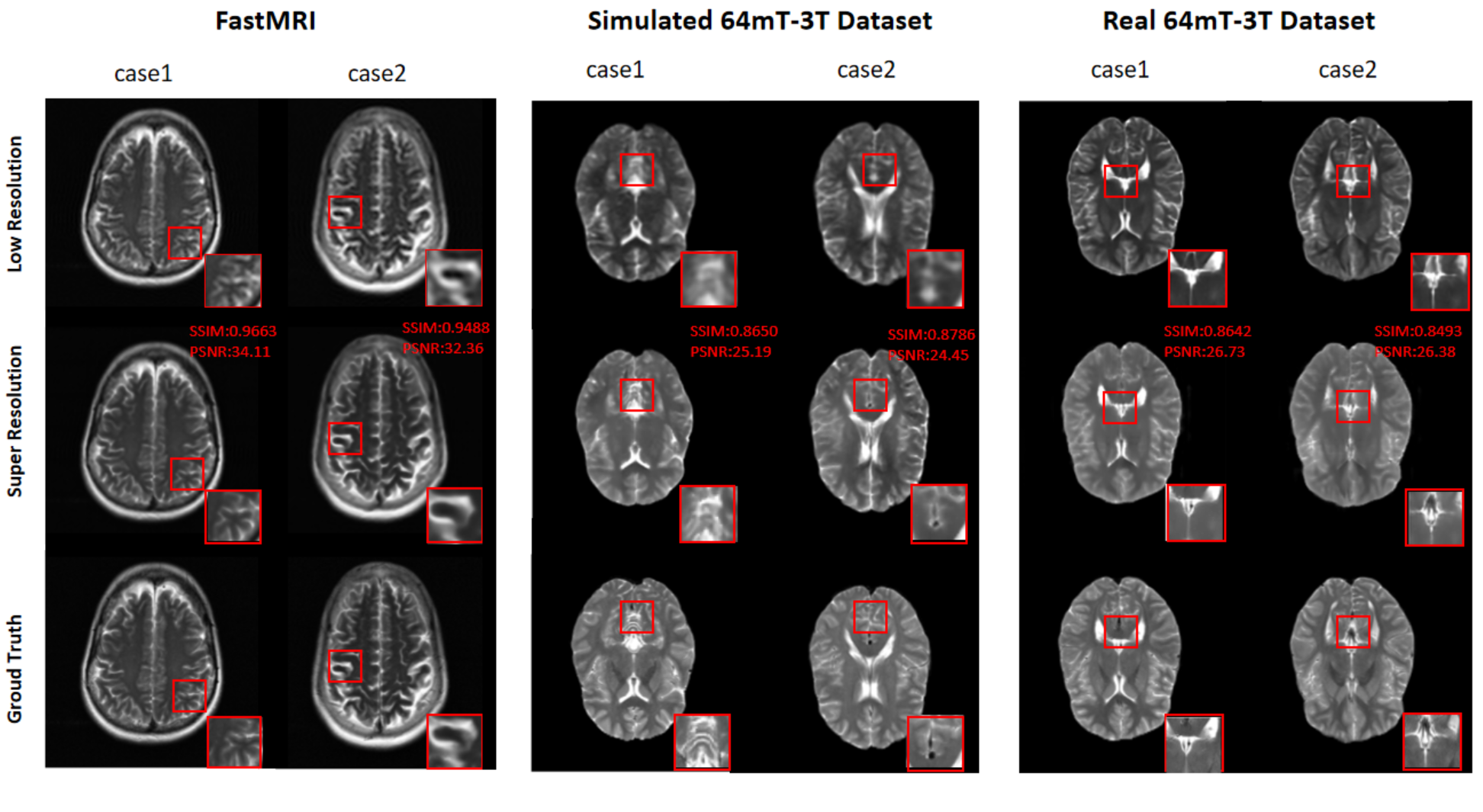}
\caption{The qualitative results on three simulated datasets under the optimal input resolution. Our method reconstructs fine details of the Ground Truth on these datasets. Red boxes highlight regions for detailed comparison.  }
\label{fig:Fig3}
\end{figure*}

\subsection{Evaluation results}
\label{sec:results}

\subsubsection{Evaluation metrics.}
Following prior work~\cite{chen2021liif,lee2022lte}, we adopt four complementary metrics for evaluation: Peak Signal-to-Noise Ratio (PSNR)~\cite{huynh2008scope} for pixel-level fidelity, Structural Similarity Index (SSIM)~\cite{wang2004image} for structural consistency, Deep Image Structure and Texture Similarity (DISTS)~\cite{ding2022dists} for perceptual quality in deep feature space and High-Frequency Error Norm (HFEN)~\cite{ravishankar2011mr} using a Laplacian of Gaussian filter for evaluating fine structural recovery.

\subsubsection{Results on dynamic resolution}

\textbf{Quantitative Results.} As shown in Table~\ref{tab:quantitative_results}, on the simulated IXI dataset, our model outperforms the three baseline methods across all resolution scales in terms of SSIM, PSNR, HFEN, and DISTS. At the $\times 0.7$ resolution scale, our method achieves a PSNR of 28.10~dB, SSIM of 0.9234, HFEN of 0.3051, and DISTS of 0.1148, representing the best performance among all scales. Among all resolution scales, we observe a non-monotonic relationship between input resolution and SR performance. This finding supports our hypothesis that pursuing the highest input resolution is not always optimal for SR. To further validate whether this observation holds in real-world scenarios, we test our method on multi-resolution tasks using the real-world 3T--5T dataset. Specifically, at $\times 0.76$, our method achieves the lowest HFEN (0.4570) and competitive performance across other metrics (SSIM: 0.8775, PSNR: 26.76~dB), indicating superior high-frequency structure recovery. Interestingly, the highest PSNR (26.78~dB) and SSIM (0.8775) are achieved at $\times 1.04$, while the lowest DISTS (0.1227) is obtained at $\times 0.83$. This suggests that different metrics may favor slightly different resolution settings, but the overall trend confirms that the optimal resolution lies in an intermediate range rather than at the extreme ends. These results strengthen our hypothesis: \textbf{the optimal resolution for MRI super-resolution is not necessarily the highest achievable resolution}.

\textbf{Qualitative Results.} As shown in Figure~\ref{fig:fig_4}, we validate our proposed Dynamic-Resolution hypothesis. At $\times 1.0$, the SR results exhibit amplified noise and artifacts. At $\times 0.25$, excessive degradation leads to loss of structural information that cannot be recovered. At the optimal resolution ($\times 0.7$), the reconstructions exhibit the sharpest edges and richest structural details most consistent with the ground truth.
As shown in Figure~\ref{fig:3T5T}, our method produces reconstructions that are more consistent with the 5T Ground Truth, exhibiting improved structural fidelity and realistic textures. The results at intermediate resolutions show better preservation of fine anatomical details compared to both the highest and lowest resolution inputs. On the simulated dataset and the real multi-resolution dataset, our experimental results both support the hypothesis that \textbf{the optimal resolution for MRI super-resolution is not necessarily the highest achievable resolution}.

\begin{figure*}[t]
\centering
\includegraphics[width=1.0\textwidth]{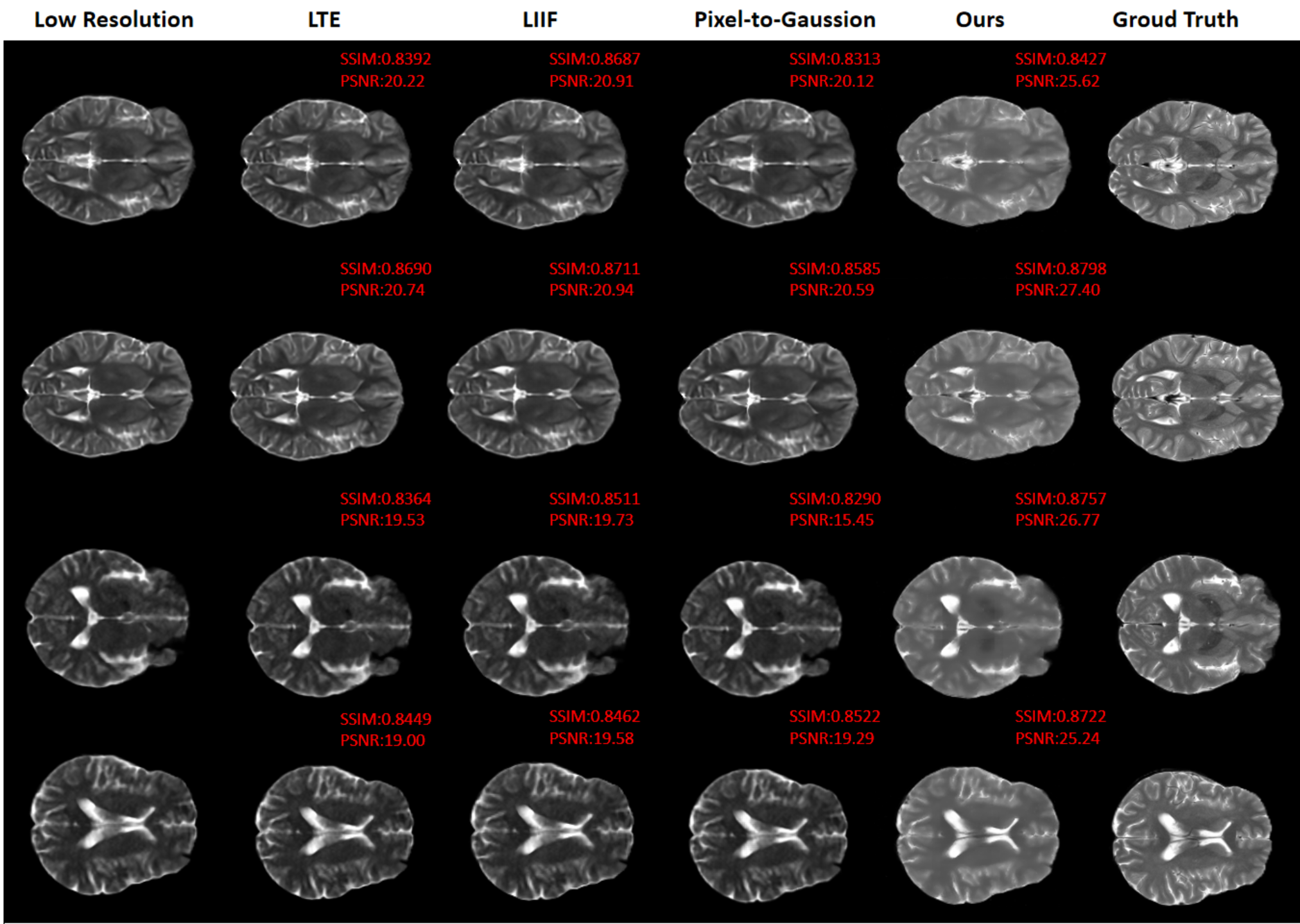}
\caption{Visual comparison on real 64mT-3T dataset. The SR results of our method can recover fine details from the Ground Truth.}
\label{fig:vis_comparsion}  
\end{figure*}

\subsubsection{Results on static resolution}

\textbf{Quantitative Results. }As shown in Table~\ref{tab:fastmri_comparison}, we quantitatively compare our method with 14 state-of-the-art approaches across three challenging SR scales ($4\times$, $5\times$, and $6.4\times$). Our method consistently achieves the best performance across all scales and metrics by a significant margin. At the $4\times$ scale, our method achieves 34.26~dB PSNR and 0.962 SSIM, surpassing the second-best method (MS-PRDDiff) by \textbf{1.45~dB} in PSNR and \textbf{0.020} in SSIM. For the more challenging $5\times$ scale, our approach attains 32.20~dB PSNR and 0.946 SSIM, with improvements of 0.30~dB and 0.032 over MS-PRDDiff, respectively. Even at the most demanding $6.4\times$ scale, our method maintains superior performance with 30.06~dB PSNR and 0.917 SSIM.  These results demonstrate substantial improvements over both traditional approaches (e.g., SwinIR, MetaSR) and recent diffusion-based methods (e.g., DiffMSR, SS-PRDDiff), validating the effectiveness and generalizability of our framework beyond ultra-low-field MRI super-resolution.

As shown in Table~\ref{tab:real_64mt}, our method with meta-learning-based domain adaptation significantly outperforms existing continuous SR approaches on real 64mT--3T data. Specifically, our method achieves 26.85~dB PSNR, substantially surpassing LIIF (20.28~dB), LTE (20.09~dB), and Pixel-to-Gaussian (19.97~dB) by more than \textbf{6.5~dB}. The higher SSIM (0.8856 vs. $<$0.862 for baselines) indicates better structural preservation. More notably, our method achieves significantly lower HFEN (0.5432 vs. $>$0.75 for baselines), demonstrating superior recovery of high-frequency anatomical details.

\textbf{Qualitative Results.} Figure~\ref{fig:Fig3} shows the qualitative results of our model on three simulated datasets under the optimal input resolution. Our method reconstructs fine details of the Ground Truth on these datasets. This is not merely an improvement in resolution, as the SR data more faithfully recovers the anatomical structures of the Ground Truth, thereby enhancing the image quality of the LR scans and improving their diagnostic utility. The visual comparisons between our mothod and baselines on the fastMRI dataset are provided in the S4.

Figure~\ref{fig:vis_comparsion} presents the qualitative results of our Static-Resolution experiments on the real 64mT--3T dataset. Baseline methods (LIIF, LTE, Pixel-to-Gaussian) produce over-smoothed results that lack fine anatomical details or introduce hallucinated structures. In contrast, our method faithfully recovers brain anatomical structures, preserving cortical folding patterns, white-gray matter boundaries, and ventricular details that are consistent with the 3T reference.


\begin{table}[t]
\centering
\caption{Quantitative comparison on the real paired 64mT--3T dataset. Our method with meta-learning-based domain adaptation significantly outperforms baseline approaches.}
\label{tab:real_64mt}
\small
\begin{tabular}{lcccc}
\toprule
Method & PSNR$\uparrow$ & SSIM$\uparrow$ & HFEN$\downarrow$ & DISTS$\downarrow$ \\
\midrule
LIIF              & 20.28 & 0.8617 & 0.8494 & 0.1645 \\
LTE               & 20.09 & 0.8596 & 0.8585 & 0.1679 \\
Pixel-to-Gaussian & 19.97 & 0.8508 & 0.7573 & 0.1726 \\
\midrule
\textbf{Ours}     & \textbf{26.85} & \textbf{0.8856} & \textbf{0.5432} & \textbf{0.1454} \\
\bottomrule
\end{tabular}
\end{table}

\subsection{Ablation Study}
\label{abla}
To verify the contribution of each proposed component, we conduct ablation experiments on the Real 64mT-3T paired dataset. 

\begin{wraptable}{r}{0.48\columnwidth}
\vspace{-4mm}
\centering
\caption{Impact of initialization.}
\vspace{-2mm}
\label{tab:init}
\footnotesize
\setlength{\tabcolsep}{4pt}
\renewcommand{\arraystretch}{0.95}
\begin{tabular}{lcc}
\toprule
Method & PSNR $\uparrow$ & SSIM $\uparrow$ \\
\midrule
Baseline  & 18.2 & 0.71 \\
Only init & 19.1 & 0.68 \\
Ours      & \textbf{25.4} & \textbf{0.85} \\
\bottomrule
\end{tabular}
\vspace{-3mm}
\end{wraptable}

\textbf{Impact of Initialization.} To validate the effectiveness of Segmentation-Guided Primitive Initialization, we compare three variants: (1) the baseline without any proposed modules, (2) the baseline with a naive initialization strategy (Only init), and (3) our full model. As shown in Table~\ref{tab:init}, our naive initialization design improves PSNR from 18.2 to 19.1, indicating better pixel-wise fidelity. However, it leads to a decrease in SSIM (from 0.71 to 0.68), suggesting that our model with initialization lacks structural consistency and may introduce anatomical distortions. These results demonstrate that while a simple initialization strategy can enhance reconstruction, it fails to preserve structural information.

\begin{wraptable}{r}{0.48\columnwidth}
\vspace{-4mm}
\centering
\caption{Impact of physics constraint.}
\vspace{-2mm}
\label{tab:physics}
\footnotesize
\setlength{\tabcolsep}{3pt}
\renewcommand{\arraystretch}{0.95}
\begin{tabular}{lcc}
\toprule
Method & PSNR $\uparrow$ & SSIM $\uparrow$ \\
\midrule
Baseline     & 18.2 & 0.71 \\
w/o Physics  & 21.6 & 0.55 \\
Ours         & \textbf{25.4} & \textbf{0.85} \\
\bottomrule
\end{tabular}
\vspace{-3mm}
\end{wraptable}
\textbf{Impact of Physics Constraint.} To evaluate the effectiveness of the proposed physics-constrained signal modeling,
we compare three variants: (1) the baseline model, (2) a model without physics-constrained modeling (w/o Physics), and (3) the full model. As shown in Table~\ref{tab:physics}, the variant without the physics constraint leads to clear performance degradation, especially in SSIM (from 0.85 to 0.55), indicating a significant loss of structural consistency. In addition, PSNR also decreases compared to the full model (from 25.4 to 21.6). These results demonstrate that the physics-constrained design plays a crucial role in improving reconstruction quality by enforcing structural consistency and suppressing unrealistic artifacts.

\begin{wraptable}{r}{0.48\columnwidth}
\vspace{-4mm}
\centering
\caption{Impact of meta-learning.}
\label{tab:meta}
\footnotesize
\setlength{\tabcolsep}{4pt}
\renewcommand{\arraystretch}{0.95}
\begin{tabular}{lcc}
\toprule
Method & PSNR $\uparrow$ & SSIM $\uparrow$ \\
\midrule
Baseline     & 18.2 & 0.71 \\
w/o Physics  & 19.7 & 0.82 \\
Ours         & \textbf{25.4} & \textbf{0.85} \\
\bottomrule
\end{tabular}
\vspace{-3mm}
\end{wraptable}
\textbf{Impact of Meta-learning.} To evaluate the effectiveness of the proposed meta-learning strategy, we compare three variants:
(1) the baseline model without any proposed components, 
(2) a model without meta-learning (w/o Meta), and (3) the full model. As shown in Table~\ref{tab:meta}, the w/o Meta variant achieves moderate improvements over the baseline in both PSNR (18.2 $\rightarrow$ 19.7) and SSIM (0.71 $\rightarrow$ 0.82), indicating that the other components contribute to better reconstruction quality. However, the full model significantly outperforms the w/o Meta variant, with PSNR increasing from 19.7 to 25.4 and SSIM from 0.82 to 0.85. By learning from diverse degradation patterns, the meta-learning strategy effectively improves the model's generalization ability, enabling our model to better handle domain variations.

The quantitative results are reported in the supplementary material S4.2. These results further validate the effectiveness of the proposed components and support the conclusions drawn from the main ablation study.

\section{Conclusion}

We reveal that the low-resolution image may be not the optimal one acquired from the MRI system due to the MRI image quality can vary substantially in both SNR and resolution. Based on this finding, we propose a hypothesis that the optimal resolution for MRI super-resolution is not necessarily the highest achievable resolution. To validate this, we propose a physics-informed MRI super-resolution framework based on 2D Gaussian Splatting. Our methods confirm the proposed hypothesis, and outperform all baseline methods on both dynamic-resolution datasets and the FastMRI benchmark in terms of quantitative metrics and qualitative comparisons.

\section{Acknowledgment}
This work was supported by the ShanghaiTech AI4S Initiative. The authors gratefully acknowledge the Advanced MR Imaging Research Laboratory, School of Biomedical Engineering, ShanghaiTech University, for providing the imaging facilities and technical support. Computational resources were provided in part by the High-Performance Computing Shared Services Platform at ShanghaiTech University.

\newpage

\bibliographystyle{IEEEtran}
\bibliography{IEEEabrv,main}

\newpage

\setcounter{section}{0}
\renewcommand{\thesection}{S\arabic{section}}
\renewcommand{\thesubsection}{S\arabic{section}.\arabic{subsection}}
\renewcommand{\thefigure}{S\arabic{figure}}
\renewcommand{\thetable}{S\arabic{table}}
\renewcommand{\theequation}{S\arabic{equation}}

\section{Theoretical Background: Resolution-SNR Trade-off in MRI}
\label{sec:supp_theory}

MRI does not directly measure image intensities at discrete pixels. Instead, the received signal is a Fourier encoding of the continuous spin distribution modulated by the imaging sequence and receiver coils. After demodulation, the signal received by the \(c\)-th coil at time \(t\) can be written as
\begin{equation}
s_c(t)
=
\int_{\Omega}
\rho(\mathbf{r})
C_c(\mathbf{r})
M(\mathbf{r};TR,TE)
e^{-i2\pi \mathbf{k}(t)\cdot \mathbf{r}}
d\mathbf{r}
+
n_c(t),
\label{eq:supp_cont_signal}
\end{equation}
where \(\mathbf{r}\) denotes spatial location, \(\rho(\mathbf{r})\) is the proton density, \(C_c(\mathbf{r})\) is the coil sensitivity, \(M(\mathbf{r};TR,TE)\) is the sequence-dependent contrast weighting, \(\mathbf{k}(t)\) is the k-space trajectory determined by the applied gradients, and \(n_c(t)\) is the acquisition noise. For a spin-echo acquisition, the contrast term can be approximated as
\begin{equation}
M(\mathbf{r};TR,TE)
\propto
B_0
\left(1-e^{-TR/T_1(\mathbf{r})}\right)
e^{-TE/T_2(\mathbf{r})},
\label{eq:supp_spin_echo}
\end{equation}
where \(B_0\) is the main magnetic field strength, and \(T_1(\mathbf{r})\) and \(T_2(\mathbf{r})\) are the longitudinal and transverse relaxation times, respectively. Substituting Eq.~\eqref{eq:supp_spin_echo} into Eq.~\eqref{eq:supp_cont_signal} gives the standard interpretation that the measured signal is proportional to the number of contributing spins, weighted by tissue contrast and coil sensitivity.

For notational simplicity, we define the contrast-weighted spin density of the \(c\)-th coil as
\begin{equation}
m_c(\mathbf{r})
=
\rho(\mathbf{r}) C_c(\mathbf{r}) M(\mathbf{r};TR,TE),
\label{eq:supp_weighted_density}
\end{equation}
Then Eq.~\eqref{eq:supp_cont_signal} becomes
\begin{equation}
s_c(t)
=
\int_{\Omega}
m_c(\mathbf{r})
e^{-i2\pi \mathbf{k}(t)\cdot \mathbf{r}}
d\mathbf{r}
+
n_c(t),
\label{eq:supp_kspace_signal}
\end{equation}
This equation shows that MRI acquisition samples the Fourier transform of \(m_c(\mathbf{r})\). Due to finite k-space coverage, the reconstructed image is not an ideal point-wise sampling of \(m_c(\mathbf{r})\), but a local average determined by the point spread function (PSF). The reconstructed value at voxel \(j\) can be expressed as
\begin{equation}
I_{c,j}
\approx
\int_{\Omega}
m_c(\mathbf{r}) h_j(\mathbf{r}) d\mathbf{r}
+
\eta_{c,j},
\label{eq:supp_psf}
\end{equation}
where \(h_j(\mathbf{r})\) is the PSF centered at voxel \(j\), and \(\eta_{c,j}\) is the corresponding image-domain noise. If the tissue properties and coil sensitivity vary slowly inside the voxel and the main lobe of \(h_j(\mathbf{r})\) is approximated by the voxel support \(\mathcal{V}_j\), Eq.~\eqref{eq:supp_psf} reduces to
\begin{equation}
S_{c,j}
\propto
\int_{\mathcal{V}_j}
m_c(\mathbf{r}) d\mathbf{r}
\approx
\bar{m}_{c,j} V_{\mathrm{vox}},
\label{eq:supp_voxel_signal}
\end{equation}
where \(\bar{m}_{c,j}\) is the average contrast-weighted spin density within the voxel and
\begin{equation}
V_{\mathrm{vox}}
=
\Delta x \Delta y \Delta z,
\label{eq:supp_voxel_volume}
\end{equation}
is the voxel volume. Therefore, under a fixed imaging protocol and within locally homogeneous tissue, the voxel signal scales linearly with the voxel volume.

\IEEEpubidadjcol

The acquisition noise is dominated by thermal noise from the receiver chain and is commonly modeled as zero-mean complex Gaussian noise. After linear Fourier reconstruction and coil combination, coil-dependent constants can be absorbed into a proportionality factor. The image-domain noise standard deviation then follows the conventional scaling
\begin{equation}
\sigma_{\mathrm{img}}
\propto
\sqrt{\frac{\mathrm{BW}}{N_{\mathrm{ex}}}},
\label{eq:supp_noise_scaling}
\end{equation}
where \(\mathrm{BW}\) is the receiver bandwidth and \(N_{\mathrm{ex}}\) is the number of excitations or averages. This relation reflects that the noise variance increases approximately linearly with receiver bandwidth, while averaging \(N_{\mathrm{ex}}\) independent acquisitions reduces the variance by a factor of \(N_{\mathrm{ex}}\). Combining Eqs.~\eqref{eq:supp_voxel_signal}--\eqref{eq:supp_noise_scaling}, the voxel-wise SNR satisfies
\begin{equation}
\mathrm{SNR}_{j}
=
\frac{S_j}{\sigma_{\mathrm{img}}}
\propto
\frac{
\bar{m}_{j}
\Delta x \Delta y \Delta z
\sqrt{N_{\mathrm{ex}}}
}{
\sqrt{\mathrm{BW}}
},
\label{eq:supp_snr_scaling}
\end{equation}
where \(\bar{m}_{j}\) denotes the coil-combined average contrast-weighted spin density within voxel \(j\). Thus, when tissue contrast, receiver bandwidth, and averaging are fixed,
\begin{equation}
\mathrm{SNR}
\propto
V_{\mathrm{vox}}
=
\Delta x \Delta y \Delta z,
\label{eq:supp_snr_voxel}
\end{equation}

This dependence directly connects SNR to spatial resolution. For Cartesian sampling, the voxel spacing is determined by the field of view (FOV) and sampling matrix:
\begin{equation}
\Delta x=\frac{\mathrm{FOV}_x}{N_x},\quad
\Delta y=\frac{\mathrm{FOV}_y}{N_y},\quad
\Delta z=\frac{\mathrm{FOV}_z}{N_z},
\label{eq:supp_fov_matrix}
\end{equation}

Equivalently, the nominal resolution is determined by the maximum sampled spatial frequency,
\begin{equation}
\Delta x \approx \frac{1}{2k_{x,\max}},\quad
\Delta y \approx \frac{1}{2k_{y,\max}},\quad
\Delta z \approx \frac{1}{2k_{z,\max}},
\label{eq:supp_kmax_resolution}
\end{equation}
Therefore, improving resolution requires extending the sampled k-space extent, 
i.e., increasing \(k_{x,\max}\), \(k_{y,\max}\), and \(k_{z,\max}\), which according to
Eq.~\eqref{eq:supp_kmax_resolution} reduces the corresponding spatial resolutions
\(\Delta x\), \(\Delta y\), and \(\Delta z\). The resulting decrease in
\(V_{\mathrm{vox}}\) in Eq.~\eqref{eq:supp_snr_voxel} consequently reduces SNR.

The resulting penalty can be quantified explicitly. Suppose the isotropic resolution is improved by a factor of \(\alpha\), i.e.,
\begin{equation}
(\Delta x',\Delta y',\Delta z')
=
\frac{1}{\alpha}
(\Delta x,\Delta y,\Delta z),
\label{eq:supp_iso_resolution}
\end{equation}
Then
\begin{equation}
V_{\mathrm{vox}}'
=
\Delta x'\Delta y'\Delta z'
=
\frac{V_{\mathrm{vox}}}{\alpha^3},
\label{eq:supp_iso_volume}
\end{equation}
Using Eq.~\eqref{eq:supp_snr_voxel}, the SNR becomes
\begin{equation}
\mathrm{SNR}'
=
\frac{\mathrm{SNR}}{\alpha^3},
\label{eq:supp_iso_snr}
\end{equation}
Since Eq.~\eqref{eq:supp_snr_scaling} also gives \(\mathrm{SNR}\propto\sqrt{N_{\mathrm{ex}}}\), restoring the original SNR by averaging alone requires
\begin{equation}
N_{\mathrm{ex}}'
=
\alpha^6 N_{\mathrm{ex}},
\label{eq:supp_iso_nex}
\end{equation}
For example, doubling the isotropic resolution reduces the voxel-wise SNR by a factor of \(8\), and recovering the original SNR would require \(64\) times more averages.

For 2D acquisitions, if only the in-plane resolution is improved while the slice thickness is fixed, i.e.,
\begin{equation}
(\Delta x',\Delta y',\Delta z')
=
\left(
\frac{\Delta x}{\alpha},
\frac{\Delta y}{\alpha},
\Delta z
\right),
\label{eq:supp_inplane_resolution}
\end{equation}
then
\begin{equation}
V_{\mathrm{vox}}'
=
\frac{V_{\mathrm{vox}}}{\alpha^2},
\qquad
\mathrm{SNR}'
=
\frac{\mathrm{SNR}}{\alpha^2},
\qquad
N_{\mathrm{ex}}'
=
\alpha^4 N_{\mathrm{ex}},
\label{eq:supp_inplane_scaling}
\end{equation}
This in-plane case has a milder scaling than isotropic 3D resolution enhancement, but still results in a substantial SNR loss.

Finally, increasing \(N_{\mathrm{ex}}\) directly increases acquisition time. For Cartesian acquisitions, the scan time approximately scales as
\begin{equation}
T_{\mathrm{scan}}
\approx
TR \cdot N_{\mathrm{phase}} \cdot N_{\mathrm{ex}}
\label{eq:supp_scan_time_2d}
\end{equation}
for 2D imaging, and
\begin{equation}
T_{\mathrm{scan}}
\approx
TR \cdot N_y \cdot N_z \cdot N_{\mathrm{ex}},
\label{eq:supp_scan_time_3d}
\end{equation}
for 3D imaging, up to acceleration, partial Fourier, and echo-train factors. Hence, directly acquiring higher-resolution MRI leads to a fundamental trade-off among spatial resolution, SNR, and acquisition time. This theoretical relationship motivates the need to identify an appropriate resolution level that balances anatomical detail and noise robustness, rather than simply pursuing the highest nominal spatial resolution.

\section{Dataset Details}
\label{sec:supp_datasets}

This section provides comprehensive details on all datasets used in this work, including preprocessing pipelines and data splits.

\subsection{IXI-based Simulated Dynamic-Resolution Dataset}
\label{sec:supp_ixi}

We use T2-weighted 3T MRI scans from the IXI dataset as high-resolution (HR) ground truth images. All volumes undergo a standardized preprocessing pipeline:
\begin{enumerate}
    \item \textbf{Skull stripping}: Brain extraction to remove non-brain tissue.
    \item \textbf{N4 bias field correction}: Intensity inhomogeneity correction.
    \item \textbf{Intensity normalization}: Rescaling to a unified intensity range.
    \item \textbf{Spatial resampling}: Resampling to a unified voxel spacing to eliminate inter-subject variability.
\end{enumerate}

For 2D slice extraction, we discard the top 20\% and bottom 20\% of axial slices to exclude low-information regions (e.g., skull base and vertex). From the remaining central region, we uniformly sample 10 slices per volume, yielding a total of 1,750 training slices. The corresponding low-resolution images are generated using the degradation model described in the following Section~\ref{sec:supp_degradation}.

\subsubsection{\textbf{Degradation Model for Low-Field MRI Simulation}}
\label{sec:supp_degradation}

To simulate low-field (64mT) MRI from high-field (3T) images, we adopt a physics-inspired degradation model that accounts for resolution loss, noise corruption, and intensity inhomogeneity:
\begin{equation}
    y = \mathcal{C}\left( \mathcal{M} \odot \left( \mathcal{B} \cdot \left( \mathcal{D}(x) + n \right) \right) \right),
    \label{eq:degradation}
\end{equation}
where $x$ denotes the HR image, $\mathcal{D}(\cdot)$ represents resolution degradation, $n$ is additive noise, $\mathcal{B}$ is a multiplicative bias field, $\mathcal{M}$ is a brain mask, and $\mathcal{C}(\cdot)$ denotes intensity clipping.

\textbf{Resolution Degradation via PSF Blurring}

Spatial resolution loss in low-field MRI is modeled by Gaussian point spread function (PSF) blurring:
\begin{equation}
    \mathcal{D}(x) = G_{\sigma} * x,
\end{equation}
where $G_{\sigma}$ is a Gaussian kernel with standard deviation $\sigma$, controlling the degree of resolution degradation.

\textbf{Noise Modeling}

MRI noise is modeled using a Rician distribution:
\begin{equation}
    y_{\text{noisy}} = \sqrt{(x + n_r)^2 + n_i^2},
\end{equation}
where $n_r, n_i \sim \mathcal{N}(0, \sigma_n^2)$ represent real and imaginary noise components, respectively.

\textbf{Bias Field Simulation}

Intensity inhomogeneity caused by B1 field variations is simulated via a multiplicative bias field:
\begin{equation}
    \mathcal{B} = 1 + \beta \cdot G_{\sigma_b}(z),
\end{equation}
where $z \sim \mathcal{N}(0,1)$, $G_{\sigma_b}$ is a Gaussian smoothing operator, $\beta$ controls the bias strength, and $\sigma_b$ determines spatial smoothness.

\textbf{Brain Masking and Intensity Constraints}

A brain mask $\mathcal{M}$ is applied to suppress background noise. The output is constrained to be non-negative and clipped to $[0,1]$:
\begin{equation}
    y = \text{clip}(y, 0, 1).
\end{equation}

\textbf{Default Parameter Settings}

Unless otherwise specified, we use: bias field strength $\beta = 0.35$, spatial smoothness $\sigma_b = 25$. The blur parameter $\sigma$ and noise level $\alpha$ are treated as task-dependent variables and varied to construct different degradation settings for dynamic-resolution simulation and meta-learning tasks.

\subsection{Real Paired Dataset (64mT--3T)}
\label{sec:supp_64mt}

For real-world validation of domain adaptation, we adopt the publicly available paired 64mT--3T MRI dataset released by Leiden University Medical Center~\cite{van_den_broek_2024_15862148}. This dataset contains paired low-field (64mT) and high-field (3T) brain MRI scans from 11 healthy subjects.

Following the same preprocessing procedure as the IXI dataset, we discard the top and bottom 20\% of slices and sample 30 slices from the central region of each volume. The data are split as follows:
\begin{itemize}
    \item \textbf{Training}: 180 paired slices (6 subjects)
    \item \textbf{Validation}: 60 paired slices (2 subjects)
    \item \textbf{Testing}: 60 paired slices (3 subjects)
\end{itemize}

\subsection{Real Dynamic-Resolution Dataset (3T--5T)}
\label{sec:supp_3t5t}

Paired data were acquired from 10 healthy subjects using a 3T United Imaging uMR 890 scanner and a 5T United Imaging uMR Jupiter scanner. For 5T acquisitions, a T2-weighted Fast Spin Echo (FSE) \cite{constable1992factors} sequence was employed with the following parameters: TR = 4500~ms, TE = 108~ms, flip angle = 110$^\circ$, slice thickness = 5~mm, FOV = $230~\text{mm} \times 200~\text{mm}$, and matrix size = $576 \times 501$. For 3T acquisitions, the T2-weighted FSE sequence was likewise used with TR = 4545~ms, TE = 120~ms, and flip angle = 145$^\circ$. To generate dynamic-resolution inputs, the readout and phase-encoding resolutions were progressively varied from $240 \times 208$ down to $120 \times 100$, yielding scans with in-plane resolution scales of 1.04, 0.90, 0.83, 0.76, 0.69, 0.55, and 0.50. The specific scanning parameters are summarized as follows:

\begin{table}[htbp]
\centering
\caption{Acquisition parameters for 3T scans with different resolution scales.}
\label{tab:supp_resolution_scales}
\begin{threeparttable}
\small
\renewcommand{\arraystretch}{1.15}
\begin{tabularx}{\columnwidth}{@{}CCC@{}}
\toprule
Resolution scale & FOV\tnote{a} (mm) & Matrix size\tnote{b} \\
\midrule
1.04  & $230 \times 200$ & $240 \times 209$ \\
0.90  & $230 \times 200$ & $208 \times 181$ \\
0.83  & $230 \times 200$ & $192 \times 167$ \\
0.76  & $230 \times 200$ & $176 \times 153$ \\
0.69  & $230 \times 200$ & $160 \times 139$ \\
0.625 & $230 \times 200$ & $144 \times 125$ \\
0.55  & $230 \times 200$ & $128 \times 111$ \\
0.50  & $230 \times 200$ & $112 \times 97$ \\
\bottomrule
\end{tabularx}
\begin{tablenotes}
\footnotesize
\item[a] FOV is reported as readout $\times$ phase-encoding.
\item[b] Matrix size is reported as $N_x \times N_y$.
\end{tablenotes}
\end{threeparttable}
\end{table}

\subsection{Supplementary Unpaired 5T Data}
\label{sec:supp_unpaired_5t}

Due to the inconvenience and high cost of paired dynamic-resolution acquisitions, we additionally collect \textbf{unpaired} 5T scans from 18 subjects using the same 5T scanner and T2-weighted FSE sequence with the same parameters as described in Section~\ref{sec:supp_3t5t}. These unpaired 5T images serve as high-resolution references, from which we synthesize pseudo-paired 3T images at multiple resolution scales using our degradation model described in Section~\ref{sec:supp_degradation}. This simulation-assisted strategy enables effective model training without requiring extensive paired acquisitions.

\section{Additional Implementation Details}
\label{sec:supp_implementation}

\subsection{MRI-Specific Covariance Dictionary Statistics}
We collect 4,241 5T MRI slices and fit each with 2D Gaussian primitives to obtain optimal representations. Statistical analysis reveals that the covariance parameters $(\sigma_x^2, \sigma_y^2, \rho\sigma_x\sigma_y)$ follow characteristic distributions within the ranges $[0, 2.39]$, $[-0.18, 1.12]$, and $[0, 2.05]$, respectively, which are notably distinct from those observed in natural images. We then sample 10 values for each parameter, yielding $10^3 = 1{,}000$ unique covariance kernel combinations. Appending the zero kernel $(0, 0, 0)$ results in a final dictionary of \textbf{1,001 covariance kernels}.

\subsection{Segmentation Mask Generation}

For high-resolution images (e.g., 3T/5T MRI), we employ the open-source pre-trained model SynthSeg~\cite{billot2023robust} to generate segmentation masks. SynthSeg is trained on synthetic MRI data and supports contrast-agnostic whole-brain segmentation, partitioning the brain into 60 anatomical regions while producing both segmentation masks and resampled volumetric data (resampled from the original voxel spacing to $1 \times 1 \times 1$~mm$^3$). We consolidate these regions into three primary tissue classes: gray matter (GM), white matter (WM), and cerebrospinal fluid (CSF). The volumetric images and their corresponding masks are subsequently divided into axial slices. Although this segmentation procedure can in principle be applied to low-field data (e.g. 64mT) to achieve uniform voxel spacing, SynthSeg fails to produce complete and reliable segmentations for such images due to their inherently low signal-to-noise ratio.

To address this limitation, we train a 2D U-Net as a dedicated low-field segmentation network using paired high-resolution masks and spatially aligned low-resolution slices. This approach enables accurate generation of three-tissue brain masks for low-resolution slices and integrates seamlessly into our 2D super-resolution framework.

\subsection{Additional details of Domain Adaptation via Meta-Learning}
\label{sec:supp_meta}

As stated before (Method Section~\ref{sec:meta}), to generalize our model to real-world low-field MRI data, we design a meta-learning-based domain adaptation framework that enables knowledge transfer from simulated degradation domains to real paired 64mT--3T data (Experiment Section~\ref{sec:results}). Here we provide more details serving as a proof-of-concept for deploying our super-resolution framework in practical scenarios where paired training data is scarce.

\subsubsection{Motivation}
\label{sec:supp_meta_motivation}

Despite careful design, simulated low-field MRI inevitably differs from real acquisitions due to unmodeled physical factors and scanner-specific characteristics. Direct application of models trained on simulated data often leads to suboptimal performance on real data.

Our meta-learning framework addresses this challenge by:
\begin{enumerate}
    \item Learning a robust initialization that can quickly adapt to new degradation domains with minimal samples.
    \item Leveraging limited real paired data as target-domain guidance during meta-training.
    \item Modeling domain variability through task-based episodic training across a spectrum of simulated and real degradation conditions.
\end{enumerate}

\subsubsection{Task Formulation}
\label{sec:supp_meta_tasks}

We formulate domain adaptation as a meta-learning problem where each \textit{task} corresponds to a distinct MRI degradation domain characterized by specific blur, noise, and resolution parameters, rather than a semantic category. This enables the model to learn domain-agnostic representations that generalize across varying imaging conditions. Specifically, we construct two categories of tasks:

\textbf{Simulated Tasks} $\mathcal{T}_{\text{sim}} = \{T^s_1, \ldots, T^M_s\}$:
Simulated tasks are generated by varying degradation parameters (blur kernels, noise levels, resolution factors) to emulate diverse MRI acquisition conditions. They are ordered by increasing Fréchet Inception Distance (FID), spanning a spectrum from low-gap to high-gap relative to the real data distribution. Concretely, we design four simulated tasks as shown in Table~\ref{tab:supp_meta_tasks}:

\begin{table}[h]
\centering
\caption{Simulated task configurations with corresponding domain gaps measured by FID.}
\label{tab:supp_meta_tasks}
\small
\begin{tabular}{ccccc}
\toprule
Task & Blur ($\sigma$) & Noise ($\alpha$) & Resolution & FID \\
\midrule
$T^s_1$ (Low-gap)  & 0.8 & 0.12 & 1.0 & 18.54 \\
$T^s_2$ (Mid-gap)  & 1.4 & 0.12 & 1.0 & 21.46 \\
$T^s_3$ (Mid-gap)  & 0.8 & 0.12 & 0.7 & 25.42 \\
$T^s_4$ (High-gap) & 1.4 & 2.0  & 1.0 & 28.38 \\
\bottomrule
\end{tabular}
\end{table}

\textbf{Real Tasks} $\mathcal{T}_{\text{real}} = \{T^r_1, \ldots, T^K_r\}$:
Due to limited availability of real paired data, we partition the real 64mT--3T training set into two subsets ($K{=}2$), each forming a real task. These tasks provide direct supervision from the target distribution and are incorporated during meta-training to continuously inject target-domain constraints and prevent overfitting to synthetic artifacts.

For each task $T_i \in \mathcal{T}_{\text{sim}} \cup \mathcal{T}_{\text{real}}$, we construct a support set $\mathcal{D}^i_{\text{sup}}$ and a query set $\mathcal{D}^i_{\text{qry}}$ sampled from \textit{different subjects}, ensuring that the adaptation and evaluation phases remain independent and promote generalization.

\subsubsection{Episodic Meta-Optimization}
\label{sec:supp_meta_optim}

We adopt a first-order Model-Agnostic Meta-Learning (MAML) framework, where the meta-objective is to find a model initialization $\theta$ that can be rapidly adapted to any task with a small number of gradient steps.

\textbf{Episode Sampling}: In each meta-training episode, we sample a task batch $\mathcal{B} = \{T_{b_1}, \ldots, T_{b_n}\}$ with the constraint that \textit{at least one task must originate from} $\mathcal{T}_{\text{real}}$. This ensures continuous exposure to realistic imaging characteristics while leveraging the diversity of simulated degradations.

\textbf{Inner-Loop Adaptation}: For each sampled task $T_i \in \mathcal{B}$, the model performs a single gradient update on the support set $\mathcal{D}^i_{\text{sup}}$. Starting from the current meta-parameters $\theta$, the task-specific adapted parameters are computed as:
\begin{equation}
    \theta'_i = \theta - \alpha \nabla_{\theta} \mathcal{L}_{\text{rec}}(\theta;\, \mathcal{D}^i_{\text{sup}}),
    \label{eq:inner_loop}
\end{equation}
where $\alpha$ is the inner-loop learning rate and $\mathcal{L}_{\text{rec}}$ is the reconstruction loss defined in the main paper (Method Section~\ref{sec:loss}). In practice, the support set consists of 2 slices drawn from one subject.

\textbf{Outer-Loop Meta-Update}: After inner-loop adaptation, the adapted parameters $\theta'_i$ are evaluated on the corresponding query set $\mathcal{D}^i_{\text{qry}}$ (2 slices from a different subject). The meta-parameters $\theta$ are then updated by aggregating gradients across all tasks in the episode:
\begin{equation}
    \theta \leftarrow \theta - \beta \sum_{T_i \in \mathcal{B}} \nabla_{\theta} \mathcal{L}_{\text{rec}}(\theta'_i;\, \mathcal{D}^i_{\text{qry}}),
    \label{eq:outer_loop}
\end{equation}
where $\beta$ is the outer-loop learning rate. We employ a first-order approximation (i.e., we do not differentiate through the inner-loop update) for computational efficiency.

The overall episodic training procedure is summarized as follows:
\begin{enumerate}[label=(\roman*)]
    \item \textbf{Episode sampling}: Sample task batch $\mathcal{B}$ with at least one real task from $\mathcal{T}_{\text{real}}$.
    \item \textbf{Inner-loop adaptation}: For each $T_i \in \mathcal{B}$, compute task-specific parameters $\theta'_i$ via Eq.~\eqref{eq:inner_loop} using the support set $\mathcal{D}^i_{\text{sup}}$.
    \item \textbf{Outer-loop optimization}: Update meta-parameters $\theta$ via Eq.~\eqref{eq:outer_loop} using query sets $\{\mathcal{D}^i_{\text{qry}}\}$.
\end{enumerate}

This dual-domain episodic strategy directly addresses the data scarcity problem inherent in real-world MRI enhancement. Simulated tasks spanning a range of FID values provide diverse degradation patterns that encourage the model to learn generalizable super-resolution priors. Meanwhile, the mandatory inclusion of real tasks in every episode anchors the meta-optimization toward the target distribution, preventing the learned initialization from drifting toward synthetic artifacts. The resulting model acquires an initialization $\theta$ from which rapid, few-shot adaptation to unseen degradation conditions---including real low-field MRI acquisitions---becomes feasible with minimal paired samples.

\section{Additional Experiment Results}
\label{sec:supp_benchmark}

\subsection{More FastMRI Benchmark Results}
\label{sec:fastmri_sup}

We provide additional visual comparisons on the fastMRI benchmark to further validate the effectiveness of our proposed method.

\begin{figure}[t]    
    \centering    
    \includegraphics[width=\linewidth]{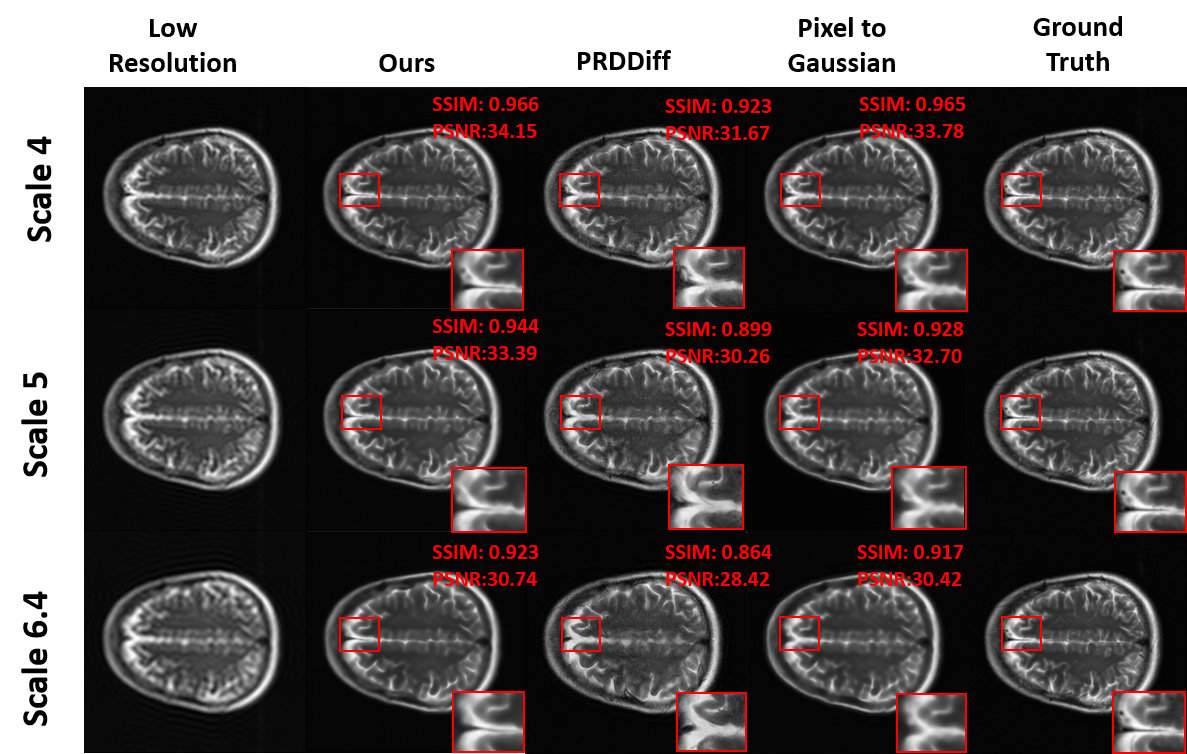}    
    \caption{Visual comparison on the fastMRI benchmark at scales $\times$4, $\times$5, and $\times$6.4. 
    Our method consistently recovers fine anatomical structures (e.g, the inter-hemispheric fissure between the left and right cerebral hemispheres) that are lost by competing methods. Pixel-to-Gaussian produces over-smoothed reconstructions, while PRDDIFF suffers from residual noise and inferior anatomical fidelity and structural consistency.}    
    \label{fig:supp_fastmri_2}
\end{figure}

Figure~\ref{fig:supp_fastmri_2} presents qualitative comparisons against state-of-the-art SR methods, including PRDDIFF and Pixel-to-Gaussian, across degradation scales of $\times$4, $\times$5, and $\times$6.4. 
Across all three scales, our method successfully recovers fine structural details that are absent in competing methods. Most notably, neither baseline is able to reconstruct the inter-hemispheric fissure separating the left and right cerebral hemispheres.
Pixel-to-Gaussian produces visibly over-smoothed outputs, losing high-frequency anatomical details and blurring tissue boundaries. 
While PRDDIFF generates sharper-looking results, it fails to faithfully recover fine-grained structures and introduces residual noise artifacts; moreover, its reconstructions exhibit weaker anatomical fidelity and structural consistency compared to our method.
These qualitative observations are fully consistent with the quantitative results reported in the main paper, further confirming the superiority of our approach in MRI super-resolution.

\subsection{More Ablation Study Results}
\label{sec:supp_ablation}

\begin{figure}[t]
    \centering
    \includegraphics[width=\linewidth]{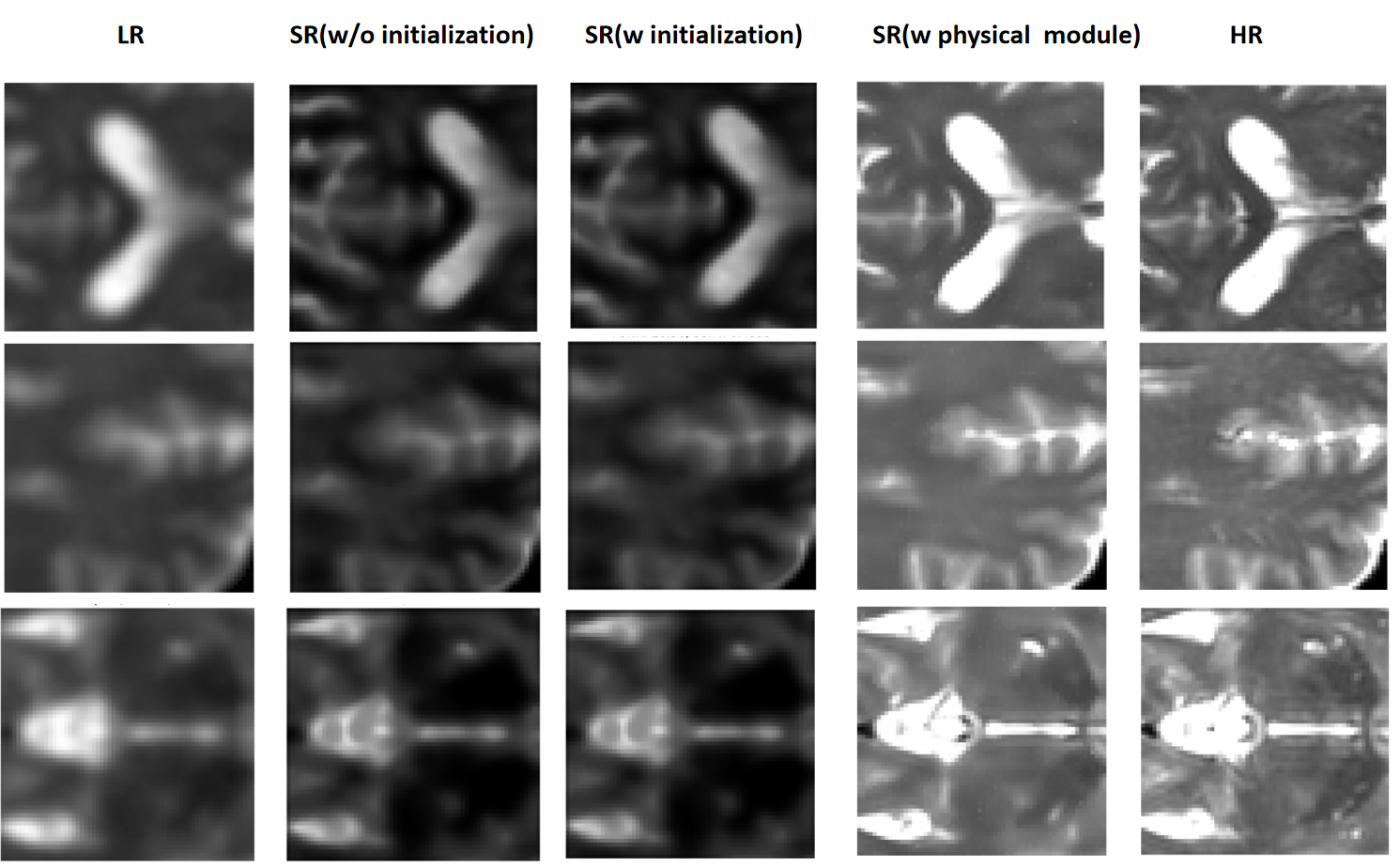}
    \caption{Visual comparison of the ablation study. 
    From left to right: LR input, w/o initialization 
    (w/o Init.), with initialization only (w/ Init.), 
    with initialization and physics module 
    (w/ Init.\ + Phy.), and ground truth (GT). 
    The incorporation of the physics module significantly 
    improves the grayscale fidelity and recovers finer 
    structural details that are consistent with the GT.}
    \label{fig:supp_ablation}
\end{figure}

Figure~\ref{fig:supp_ablation} provides additional visual 
comparisons to complement the ablation study presented in 
the main paper. We compare four configurations from left 
to right: the LR input, the result without initialization 
(w/o Init.), the result with initialization only (w/ Init.), 
the result with both initialization and the physics module 
(w/ Init. + Phy.), and the ground truth (GT).

As can be observed, before the physics module is introduced, 
the super-resolved outputs tend to retain intensity 
distributions that closely resemble the LR input, leading 
to a pronounced deviation from the GT in terms of overall 
grayscale fidelity. 
Specifically, removing the initialization module results 
in visibly blurred reconstructions, where fine structural 
details are largely smoothed out. 
Incorporating the initialization module partially alleviates 
this issue by improving the sharpness of internal structures; 
however, a noticeable gap in the global intensity distribution 
persists relative to the GT.
Upon integrating the physics module, the grayscale 
distribution is faithfully restored to be consistent with 
the GT. Moreover, finer and more anatomically accurate 
structural details are recovered, which are otherwise absent 
in the physics-free variants. These observations are 
consistent with the quantitative results reported in the 
main paper and further confirm the necessity of each 
proposed component.

\section{Ethics Statement}
\label{sec:ethic}
The dynamic-resolution 3T-5T paird data used in this study were collected under the oversight of the Ethics Committee of ShanghaiTech University (Application No. Q2026-015). Written informed consent was obtained from all participants prior to data acquisition. All data were anonymized before analysis and used only for research purposes.

The following page provides the original ethics approval document for this study.

\begin{figure*}[t]
    \centering
    \includegraphics[
        width=\textwidth,
        height=0.9\textheight,
        keepaspectratio
    ]{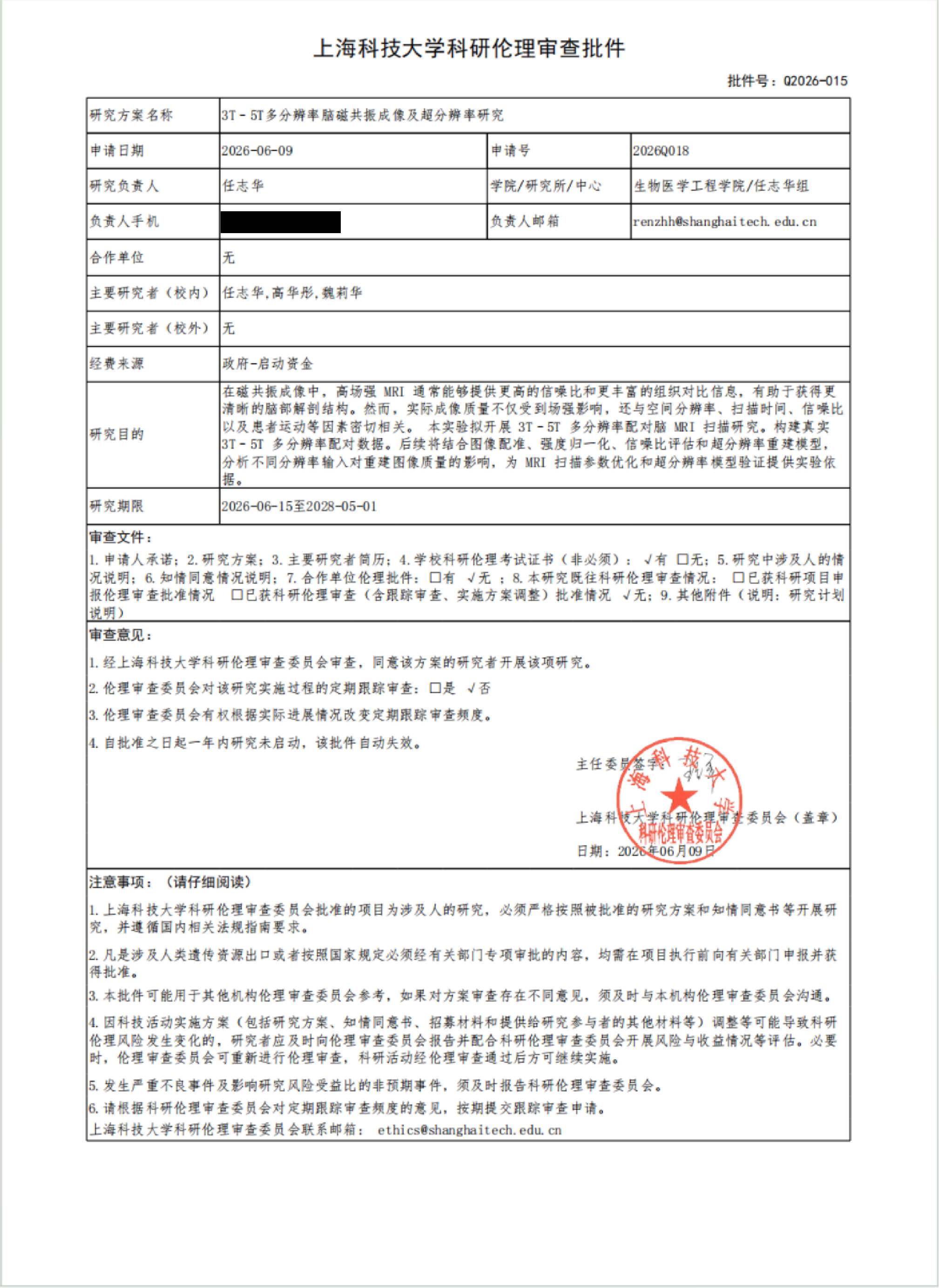}
\end{figure*}

\newpage

\section{Biography Section}
\begin{IEEEbiography}[{\includegraphics[width=1\textwidth]{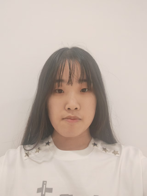}}]
{\textbf{Lihua Wei}} received the B.E. degree in Computer Science and Technology from Donghua University, China, in 2025. She is currently pursuing the M.E. degree with the School of Biomedical Engineering, Shanghaitech University, China. Her research interests include MRI super-resoluton, MRI Image enhancement, MRI image synthesis.
\end{IEEEbiography}

\begin{IEEEbiography}[{\includegraphics[width=1\textwidth]{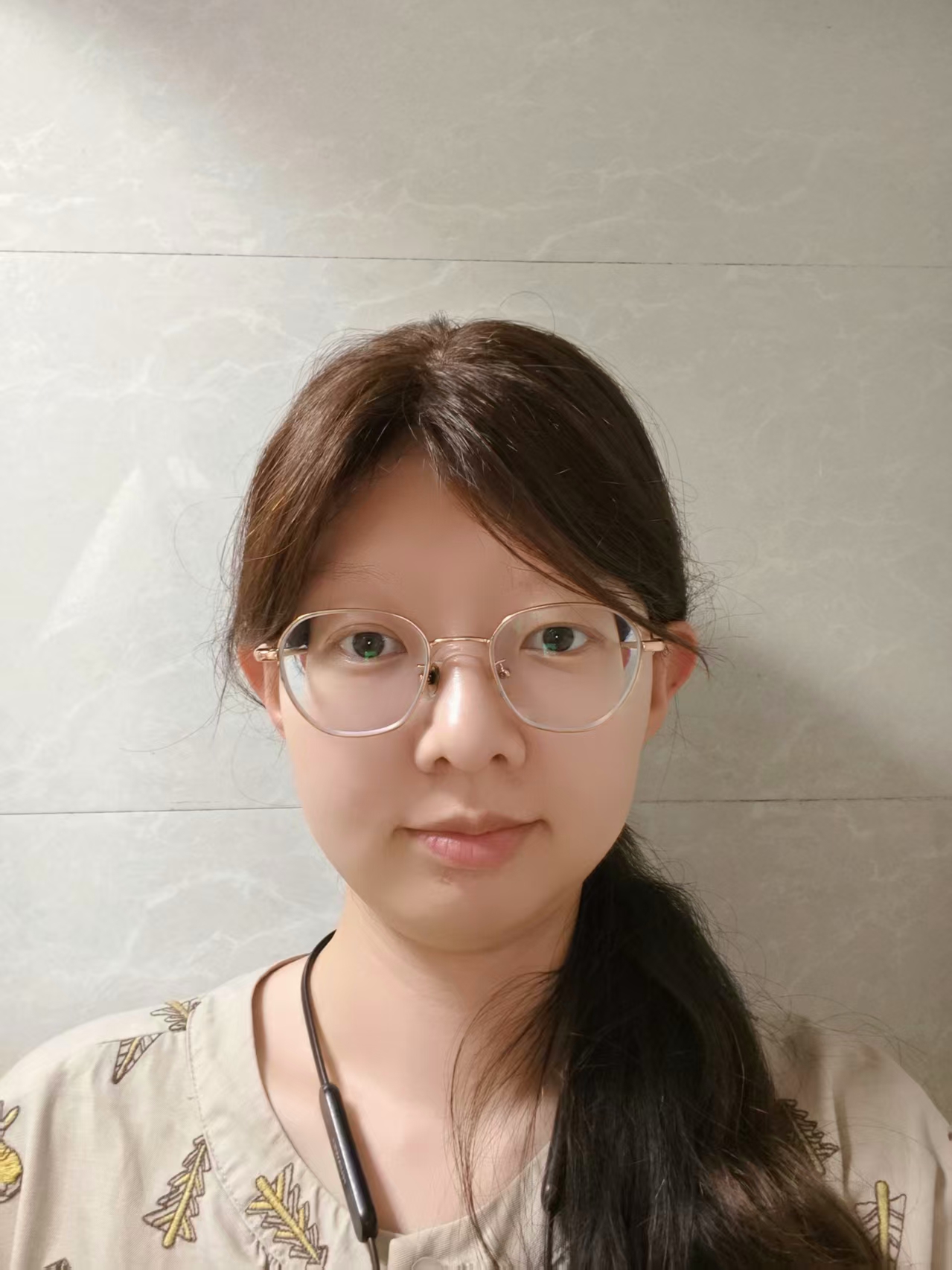}}]
{\textbf{Huatong Gao}} is a master’s student in the School of Biomedical Engineering at ShanghaiTech University, China. She received her B.Eng. degree in Biomedical Engineering from ShanghaiTech University in 2025 and is currently pursuing an M.Eng. degree in the same school. Her research interests include artificial intelligence for medical imaging, MRI super-resolution and reconstruction.
\end{IEEEbiography}

\begin{IEEEbiography}[{\includegraphics[width=1\textwidth]{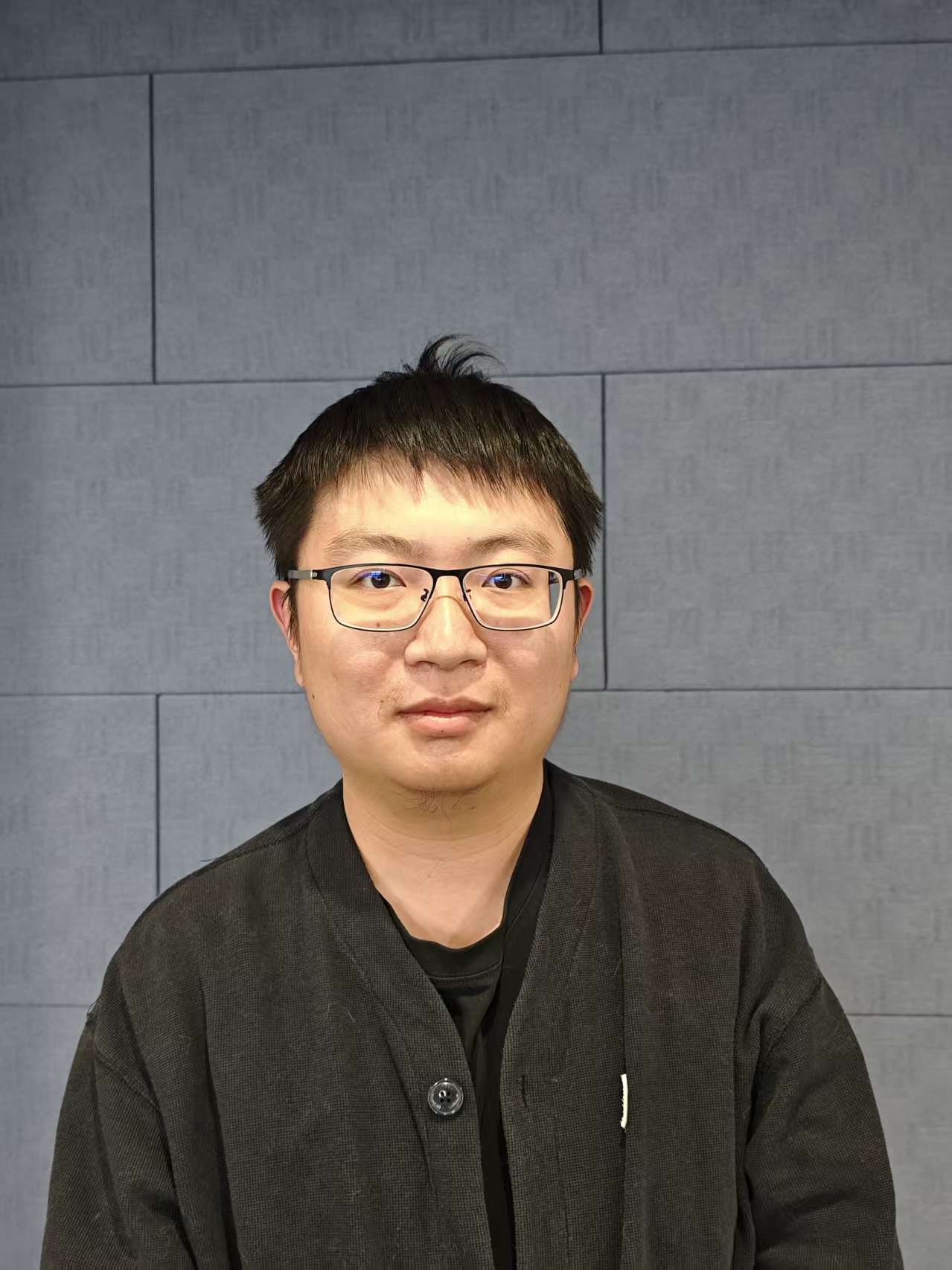}}]
{\textbf{Jia Gong}} is a researcher at the Shanghai Academy of Artificial Intelligence for Science. He received his B.Eng. degree in Optoelectronic Engineering from Chongqing University, China, and his Ph.D. in Information Systems Technology and Design from the Singapore University of Technology and Design. His research interests span human pose estimation, human mesh reconstruction, digital avatars, and active learning.
\end{IEEEbiography}

\begin{IEEEbiography}[{\includegraphics[width=1\textwidth]{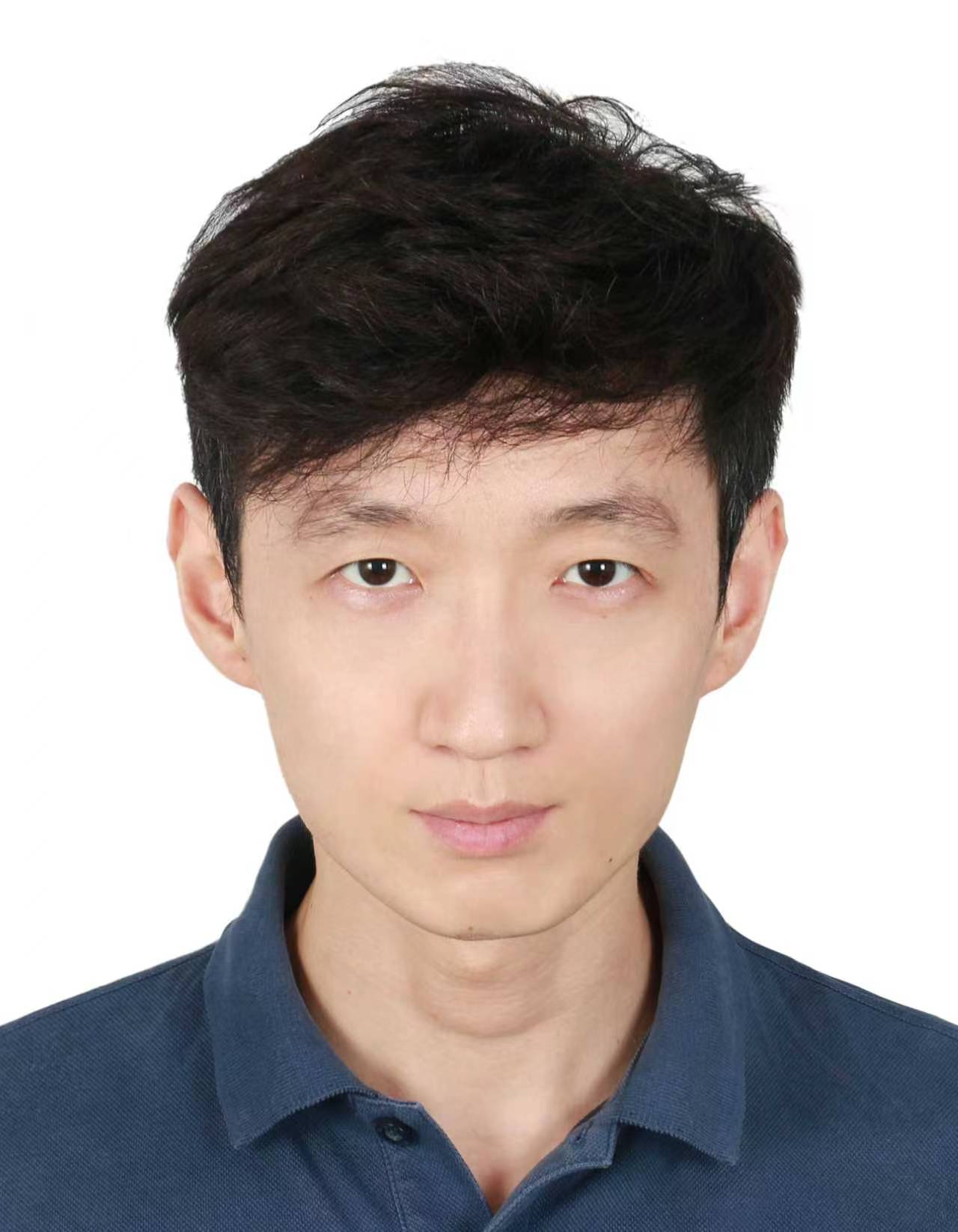}}]
{\textbf{Zhiyu Tan}} is currently pursuing the Ph.D. degree with the Artificial Intelligence Innovation and Incubation (AI³) Institute, Fudan University. His research interests include multimodal understanding and generation, reinforcement learning-based post-training methods, and multimodal large language models. His current research focuses on developing AI models with stronger visual perception, generative capability, and alignment with human instructions and preferences.
\end{IEEEbiography}

\begin{IEEEbiography}[{\includegraphics[width=1\textwidth]{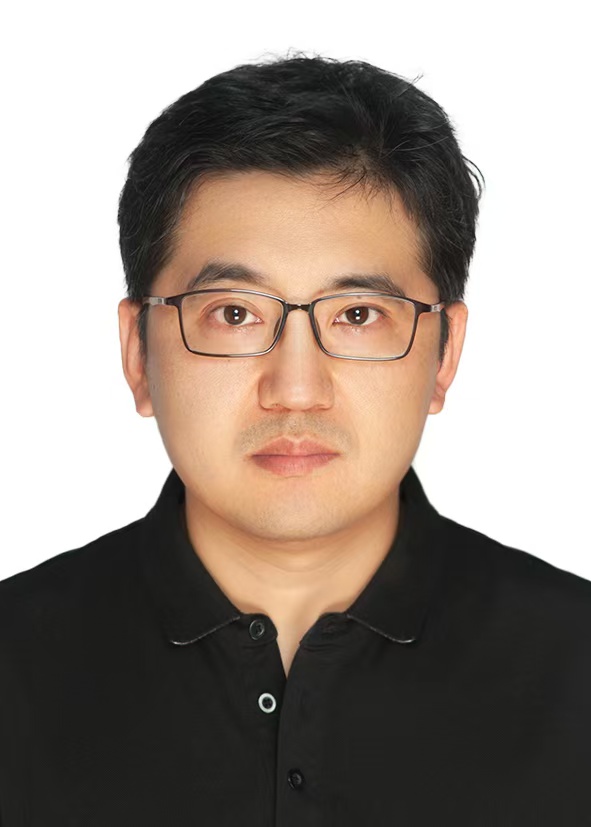}}]
{\textbf{Hao Li}} is currently a Professor at the Artificial Intelligence Innovation and Incubation Institute, Fudan University. He received his Ph.D. in 2012 from the Chinese Academy of Sciences. His research interests span video generation and AI for Science. He has published over 80 papers in top-tier venues, including CVPR, ICLR, Nature Communications, and Science Advances. He is the lead researcher of the FuXI model, a highly influential weather forecasting system.
\end{IEEEbiography}

\begin{IEEEbiography}[{\includegraphics[width=1\textwidth]{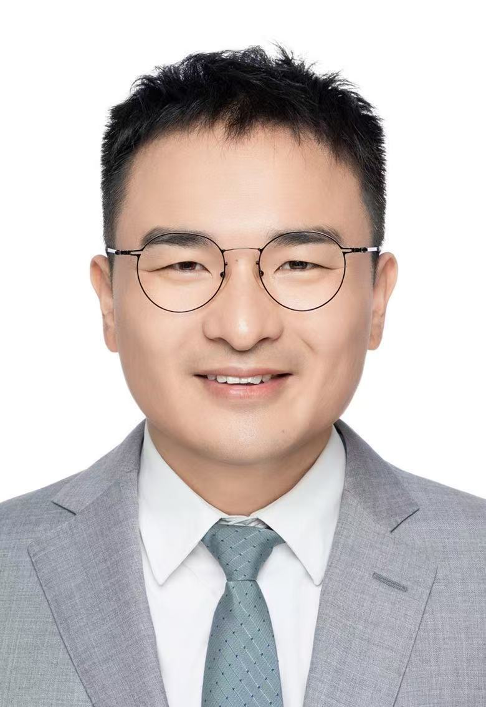}}]
{\textbf{Jun Liu}} is Professor (Chair of Digital Health) at the School of Computing and Communications, Lancaster University. He obtained his PhD from Nanyang Technological University in 2019, subsequently serving as faculty at Singapore University of Technology and Design from 2019 to 2024. Prior to his academic career, he worked at Tencent from 2014 to 2015. He is a Fellow of British Computer Society (BCS). He obtained the Best Paper Awards from PREMIA (2016, 2019), the Best Doctoral Thesis Award from EEE at NTU (2020), and the IEEE VSPC Rising Star Award (runner-up) (2024). He was nominated for the Singapore President’s Young Scientist Award in 2024. He serves as Associate Editor-in-Chief for Pattern Recognition, Senior Area Editor for IEEE Transactions on Image Processing, and Associate Editor for IEEE TCSVT, IEEE TNNLS and ACM CSUR. He received the Best Associate Editor Award of IEEE TBIOM in 2026. He is General Chair of BMVC 2026 and Program Chair of BMVC 2025, and has served as Area Chair for premier conferences including CVPR, ECCV, ICML, NeurIPS, ICLR, IJCAI, AAAI, WACV, and ACM Multimedia. His research focuses on computer vision, machine learning, and digital health applications.
\end{IEEEbiography}

\begin{IEEEbiography}[{\includegraphics[width=1\textwidth]{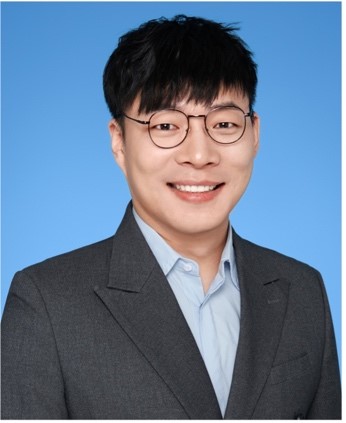}}]
{\textbf{Zhihua Ren}} is an Assistant Professor and Director of the Biomedical Imaging and Instrumentation Laboratory (Bio-Med I² Lab) at the School of Biomedical Engineering, ShanghaiTech University. He received his Ph.D. from the Singapore University of Technology and Design in 2020 and was a visiting scholar at Columbia University in 2019. He subsequently worked as a Postdoctoral Research Scientist at Columbia University for three years. His research interests include MRI physics and instrumentation, electromagnetic biomedical devices, and their applications in healthcare and medical imaging.
\end{IEEEbiography}

\vfill

\end{document}